\documentclass[10pt]{article}
\usepackage{rsi-report}
\usepackage{placeins}
\usepackage{needspace,float}
\usepackage{fontawesome5}
\definecolor{rsiLink}{HTML}{B95636}
\hypersetup{pdftitle={RSI Arena: Towards Recursive Self-Improvement with Human Feedback},pdfauthor={Yue Huang, Yanbo Wang, Lichi Li, Yuanyuan Chen, Andreas Haupt, Xiaonan Luo, Zheyuan Liu, Jihye Choi, Felix Juefei-Xu, Xiangliang Zhang, Radha Poovendran, Misha Sra, Ying Liu, Ben Burtenshaw, Thomas Wolf, Alex Pentland, Zichen Chen, Zhangchen Xu},pdfsubject={Autonomous AI research agents, post-training, human evaluation}}

\newcommand{\GapGainb}{14}

\newcommand{\GapGaind}{172}

\newcommand{\GapGainf}{13}

\newcommand{\GapGaini}{68}
\newcommand{\GapGainj}{98}

\newcommand{\GapOnea}{+1}

\newcommand{\GapOned}{−207}

\newcommand{\GapTwoa}{+7}

\newcommand{\GapTwod}{−35}

\newcommand{\GtwoTokens}{0.39}

\newcommand{\ProgAboveTwoUp}{Eight}
\newcommand{\ProgMedChange}{+2}

\newcommand{\ProgMoved}{4}

\newcommand{\ProgMovei}{18}

\newcommand{\ProgOnea}{76}

\newcommand{\ProgOnei}{57}

\newcommand{\ProgTwoa}{87}

\newcommand{\ProgTwoi}{39}

\newcommand{\ProgUpUp}{Six}
\newcommand{\RhoMoves}{−0.19}
\newcommand{\RhoProgTwo}{0.30}
\newcommand{\RhoProgTwoPlain}{0.19}
\newcommand{\RhoWildTwo}{0.79}

\newcommand{\RoundADa}{10}
\newcommand{\RoundADd}{18}
\newcommand{\RoundAK}{65}
\newcommand{\RoundAKa}{18}
\newcommand{\RoundAKk}{39}

\newcommand{\RoundDKd}{6}
\newcommand{\RoundDKk}{17}
\newcommand{\RoundSearch}{41}

\newcommand{\RoundVotes}{121}

\newcommand{\StwoClean}{1,319}

\newcommand{\StwoClosedUp}{Eight}

\newcommand{\StwoMedGain}{28}

\newcommand{\StwoPak}{58}

\newcommand{\StwoPkdPlain}{99.5}
\newcommand{\StwoPlaina}{1,041}

\newcommand{\StwoPlaind}{1,053}

\newcommand{\StwoPlaink}{1,134}
\newcommand{\StwoSCa}{1,061}

\newcommand{\StwoSCc}{1,008}
\newcommand{\StwoSCd}{1,019}

\newcommand{\StwoSCi}{1,030}

\newcommand{\StwoSCk}{1,054}

\newcommand{\StwoTestPass}{161}
\newcommand{\StwoTests}{167}

\newcommand{\StwoUsed}{1,096}
\newcommand{\StwoVoted}{1,444}

\newcommand{\TokOnea}{519}

\newcommand{\TokOned}{61}

\newcommand{\TokOnej}{326}

\newcommand{\TokTwoa}{305}

\newcommand{\TokTwod}{508}

\newcommand{\TokTwoj}{322}

\title{RSI Arena: Towards Recursive Self-Improvement\\with Human Feedback}
\author{Yue Huang\textsuperscript{1,2}\quad Yanbo Wang\textsuperscript{3}\quad Lichi Li\textsuperscript{1}\quad
  Yuanyuan Chen\textsuperscript{1}\quad Andreas Haupt\textsuperscript{4}\quad Xiaonan Luo\textsuperscript{2}\\[1pt]
  Zheyuan Liu\textsuperscript{2}\quad Jihye Choi\textsuperscript{5}\quad Felix Juefei-Xu\textsuperscript{6}\quad
  Xiangliang Zhang\textsuperscript{2}\quad Radha Poovendran\textsuperscript{7}\quad Misha Sra\textsuperscript{8}\\[1pt]
  Ying Liu\textsuperscript{9}\quad Ben Burtenshaw\textsuperscript{10}\quad Thomas Wolf\textsuperscript{10}\quad
  Alex Pentland\textsuperscript{4,11}\quad Zichen Chen\textsuperscript{1,4}\quad Zhangchen Xu\textsuperscript{1,7}}
\affiliations{%
  \affiliationentry{1}{bake-ai}{Bake AI}\hspace{9pt}
  \affiliationentry{2}{notre-dame}{University of Notre Dame}\hspace{9pt}
  \affiliationentry{3}{ucla}{University of California, Los Angeles}\hspace{9pt}
  \affiliationentry{4}{stanford}{Stanford University}\hspace{9pt}
  \affiliationentry{5}{nvidia}{NVIDIA Research}\hspace{9pt}
  \affiliationentry{6}{nyu}{New York University}\hspace{9pt}
  \affiliationentry{7}{washington}{University of Washington}\hspace{9pt}
  \affiliationentry{8}{ucsb}{University of California, Santa Barbara}\hspace{9pt}
  \affiliationentry{9}{scale}{Scale AI}\hspace{9pt}
  \affiliationentry{10}{huggingface}{Hugging Face}\hspace{9pt}
  \affiliationentry{11}{mit}{Massachusetts Institute of Technology}}
\date{}
\begin{document}
\maketitle
\begin{abstract}
Can AI agents autonomously train useful language models and improve them using feedback from users? We introduce RSI Arena, a two-stage study of autonomous post-training as a step toward recursive self-improvement. Ten frontier models act as training agents, each starting from NVIDIA Nemotron 3.5 Lightning 30B-A3B Base. During a six-day first stage, each agent has 1,000 GPU-hours and \$300 in API credit on a shared 64-GPU cluster, independently selecting data, implementing training procedures, and choosing checkpoints. We evaluate their models on 20 benchmarks and through blind human evaluations. All ten agents produce working assistants, with the strongest model, trained by GPT-6 Astra, closing a median of 76\% of the base-to-Instruct gap across 12 capability benchmarks. Its human-preference rating is also close to NVIDIA Instruct after statistical adjustment for answer length and formatting. In Stage 2, each agent receives the comparison records and votes involving its Stage 1 model and is given three additional days for further training. Using NVIDIA Instruct as a common reference across the two evaluation rounds, we observe an improvement from Stage 1 to Stage 2 in the style-adjusted relative rating of eight of the ten models. We show that benchmark gains were modest, with no clear link to preference gains, while some agents encouraged longer responses and one model showed increased harmful compliance on StrongREJECT.
RSI Arena provides an auditable setting for studying how human feedback can guide autonomous model development. It takes a step toward recursive self-improvement with human feedback (\textbf{RSIHF}), in which people help define what improvement means for the AI systems they use.

\par\vspace{8pt}
{%
  \centering
  \normalfont\sffamily
  \fontsize{9.5}{12}\selectfont
  \mbox{%
    \href{https://rsiarena.org}{%
      \textcolor{rsiLink}{%
        \faGlobe\hspace{0.45em}Project Website}}%
    \hspace{1.2em}%
    \href{https://rsiarena.org/\#b-news}{%
      \textcolor{rsiLink}{%
        \faNewspaper\hspace{0.45em}Daily Reports (Stage 1)}}%
    \hspace{1.2em}%
    \href{https://rsiarena.live}{%
      \textcolor{rsiLink}{%
        \faTv\hspace{0.45em}Training Livestream}}%
    \hspace{1.2em}%
    \href{https://huggingface.co/RSIArena}{%
      \raisebox{-1.5pt}{%
        \includegraphics[height=10pt]{%
          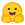}}%
      \hspace{0.45em}%
      \textcolor{rsiLink}{Model Checkpoints}}%
  }%
  \par
}

\end{abstract}
{\vspace*{-0.3em}\noindent\footnotesize\color{BakeGray}Stage 1 ran from September 29 to October 5, 2026, and Stage 2 from October 10 to 13, 2026 (Pacific Time).\par}
\vspace{0.6em}

\section{Introduction}
Recursive self-improvement is a long-term vision in which AI systems develop more capable successors that can, in turn, advance further AI development \citep{good1966,schmidhuber2007godel}. If successful, this process could accelerate AI progress beyond the pace of human research alone. Leading AI developers are actively pursuing this vision: OpenAI aims for an automated AI researcher by 2028 \citep{bellan2025openai}, and Anthropic reports that its models already lead about a quarter of its AI research and development \citep{favaro2026pace}. A practical step toward this vision is to establish whether AI agents can develop models end to end under realistic resource constraints and evaluation requirements.

AI agents already perform parts of this work. They train models for Kaggle competitions \citep{chan2024mlebench}, tackle expert-designed research engineering tasks \citep{wijk2024rebench,huang2024mlagentbench}, replicate published papers \citep{starace2025paperbench}, and write papers of their own \citep{lu2024aiscientist}. Most closely related to our work, PostTrainBench asks agents to post-train small base models, with each run targeting a single benchmark disclosed to the agent \citep{rank2026posttrainbench}. Most of these tasks take hours rather than days, use one or a few GPUs, and rely on automated evaluation. None, however, asks an agent to build a model that people will then use.

Compared with optimizing for a single benchmark, post-training a language model for real-world use poses a broader challenge: turning a pretrained model into an assistant that people find useful. Starting from a base model, a team selects data, writes or adapts training code, runs supervised fine-tuning and preference optimization, and evaluates the results \citep{ouyang2022instructgpt,lambert2024tulu3}. This process requires substantial GPU resources and repeated cycles of training and evaluation, and success cannot be captured by a single metric. Because these models are built to serve people, the ultimate test is whether users prefer them. Benchmarks assess specific capabilities through fixed response formats; to capture user preferences more directly, we also ask people to compare the models in a blind Human Arena modeled on Chatbot Arena \citep{chiang2024chatbotarena}.

RSI Arena delegates this entire process to AI agents. No human selects the training data or writes the training code; people only evaluate the resulting models. The arena consists of two stages that mirror how models are developed in practice (Figure~\ref{fig:overview}). Stage 1 captures pre-release post-training. Ten frontier models, acting as autonomous research agents, each post-trained the same base model for six days with the same budget on a shared cluster. The agents chose their own data, training code, and internal evaluations, while we livestreamed their work. We then evaluated the ten final models on 20 benchmarks and in a blind human evaluation, where participants compared pairs of model responses side by side at COLM 2026 and online. Stage 2 captures post-release improvement, when developers use user ratings and preferences to train the next version. Each agent received the arena battles involving its own model and had three additional days to improve it. This stage tests another part of the self-improvement loop: whether agents can use feedback from their models’ users to drive further improvement.

\begin{figure}[t]
\centering
\resizebox{\linewidth}{!}{\input{figures/flow.tex}}
\caption{The two stages of RSI Arena. In Stage 1 (steps 1--4), ten AI agents post-train the same base model for 144 hours, nominate checkpoints, and their final models are evaluated on 20 benchmarks and by people in the Human Arena. In Stage 2 (steps 5--8), each agent continues from its own Stage 1 model with the arena battles that model played, and the new models are evaluated again, with NVIDIA Instruct as the common reference.}
\label{fig:overview}
\end{figure}

This report covers both stages. Our main findings from Stage 1 are as follows:
\begin{enumerate}[leftmargin=1.6em, itemsep=3pt]
\item \textbf{Every agent produced a working assistant, and with style control people rate the best one level with NVIDIA Instruct.} The best agent model, trained by GPT-6 Astra, closed a median of 76\% of the base-to-Instruct gap across 12 benchmarks. Without style control, people rated NVIDIA Instruct highest (1,182, against 1,161); with answer length and formatting controlled, the two are level at 1,095 and 1,094.
\item \textbf{The agents used similar recipes.} All fine-tuned on public instruction data, and six then ran preference optimization. Six trained LoRA adapters and four trained all weights. Memory errors, storage quotas, job limits and queues took much of their time.
\item \textbf{Budget use did not predict results.} Gemini 3.8 Flash stopped after 12 hours and 19 GPU-hours. People rated its model above the models of Muse Spark 1.3 and DeepSeek V4.1 Flash, which used about 1,000 GPU-hours each.
\item \textbf{Benchmarks and people agreed only in part.} Benchmark progress and human ratings correlate at $\rho=0.73$. LLM-judged chat benchmarks track people best, WildBench at $\rho=0.95$, and competition mathematics does not. The longer answer won 72\% of decisive votes. The model with the best AIME 2026 score placed eighth of ten in human evaluation.
\item \textbf{Safety was not part of the task, and some models regressed.} Three models gave more harmful help than the base model on StrongREJECT.
\end{enumerate}
From Stage 2 we found:
\begin{enumerate}[leftmargin=1.6em, itemsep=3pt]
\item \textbf{Feedback from people helped most agents.} \StwoClosedUp{} of the ten agents narrowed their model's style-controlled gap to NVIDIA Instruct; across all ten, the median gain was \StwoMedGain{} points. The model trained by DeepSeek V4.1 Flash gained most, from \GapOned{} to \GapTwod{} points, as its median answer grew from \TokOned{} to \TokTwod{} tokens.
\item \textbf{NVIDIA Instruct is still first without style control.} It is ahead of the second model, the DeepSeek-trained one, in \StwoPkdPlain\% of bootstrap resamples, and in a targeted round it won \RoundAKk{} of its \RoundAK{} battles with the Astra-trained model, which won \RoundAKa{}. With style control, the two are level (\StwoSCa{} and \StwoSCk{}).
\item \textbf{The agents left much of their budget unused.} They used 2,880 of their 5,051 GPU-hours and \$541 of their \$1,500 of API credit; Gemini 3.8 Flash used 10 GPU-hours.
\item \textbf{Benchmarks changed little.} \ProgUpUp{} of the ten agents raised their median benchmark progress, and the median change was \ProgMedChange{} points of the gap. The changes on benchmarks and in human ratings are unrelated ($\rho=\RhoMoves$). The model trained by MiniMax M3.1 Flash lost \ProgMovei{} points, mostly because many of its answers no longer end in the format the scorers read, and it now gives more harmful help on StrongREJECT (24.5, against 2.6).
\end{enumerate}

Our goal is to connect autonomous model development with the needs and preferences of the people these models serve. RSI Arena brings users' judgments into both evaluation and subsequent development, providing an auditable setting for studying progress toward recursive self-improvement with human feedback (RSIHF). To support further study, we release the model checkpoints submitted by the agents throughout development on Hugging Face, providing an open resource for examining how model capabilities and behavior evolve during autonomous post-training.

\section{RSI Arena}
\label{sec:arena}
RSI Arena evaluates whether AI agents can autonomously develop and improve language models for real-world use. The experiment follows a two-stage development cycle: agents first post-train the same base model under matched resource budgets, then refine their models using feedback from human users. Throughout this process, agents control data selection, training, and internal evaluation, while we assess the resulting models through benchmarks and blind human comparisons.

\subsection{Task and rules}
All agents received the same instruction: to train the strongest model possible from the provided base checkpoint within their allocated resources. The instruction stated that final models would be evaluated on held-out data inaccessible to the agents and by human evaluators. The instruction left data selection and training methods to the agents and specified no target benchmark, a design choice intended to reduce incentives for reward hacking \citep{huang2026rewardhacking}. The rules prohibited training on evaluation or benchmark test data, copying reference answers, hard-coding outputs, and using other participants' checkpoints, data, or code. Any submission found to involve cheating would receive a score of zero. The instruction did not mention safety. Appendix~\ref{app:instruction} reproduces the task objective.

\paragraph{Base model}
All agents started from NVIDIA Nemotron 3.5 Lightning 30B-A3B Base \citep{nemotron35}, a mixture-of-experts model with about 30 billion parameters, of which about 3 billion are active per token. NVIDIA also released an instruct version of the same model, post-trained by NVIDIA, which we use as a reference and call NVIDIA Instruct. The base model's end-of-turn token was never trained, so the base model does not end its answers. Every agent had to teach its model to stop.

\paragraph{Budgets}
Each agent had 1,000 GPU-hours and \$300 of API credit for its own model calls. It could run up to 8 GPUs at a time, plus up to 8 preemptible GPUs that the scheduler could take back, with at most 24 hours per job and a 400\,GiB storage quota. Stage 1 lasted 144 hours. Two agents joined late, at hours 48 and 53, with the full budget. Appendix~\ref{app:cost} gives the cost of the whole study.

\paragraph{Nominations}
An agent submits a model by nominating a checkpoint. A nomination names the checkpoint directory and the completed GPU job that trained it, with a record of the training data. The platform checks the record, the files and the serving settings, then copies the checkpoint to the model registry. The agent can nominate as often as it likes. When the session ends, the latest accepted nomination becomes its final model. A session ends when the agent declares its work finished, when its budget runs out, or at the deadline.

\subsection{Agents}
Ten frontier models participated: GPT-6 Astra (OpenAI) \citep{openai2026astra}, MiMo-V2.6-Pro (Xiaomi) \citep{xiaomi2026mimov26}, Grok 4.7 (SpaceXSI) \citep{xai2026grok47}, DeepSeek V4.1 Flash (DeepSeek) \citep{deepseek2026v41flash}, GLM-5.3 (Z.ai) \citep{zai2026glm53}, Kimi K3 (Moonshot AI) \citep{kimi2026k3}, Gemini 3.8 Flash (Google) \citep{google2026gemini38flash}, Muse Spark 1.3 (Meta) \citep{meta2026musespark13}, MiniMax M3.1 Flash (MiniMax) \citep{minimax2026m31flash}\footnote{We used a preview checkpoint of MiniMax M3.1 Flash during the livestreamed experiment.}, and Step 5 Preview (StepFun) \citep{stepfun2026step5preview}. MiniMax M3.1 Flash and Step 5 Preview joined at hours 48 and 53, respectively. We refer to agents by their underlying model names and to the models they trained using shortened labels, such as the Astra-trained model and the DeepSeek-trained model. Where the context is clear, we use simply ``the model.''

All agents ran in the same harness built on the Codex agent loop \citep{codex}. Their tools run commands on their virtual machine and the cluster, read and write files, wait for jobs, report the remaining budget, nominate checkpoints and end the session. Model calls went through OpenRouter or the provider's own API. When an agent's context filled up, the harness compacted it into a summary. A session also restarted when a provider failed or we updated the platform, and the ten agents used 57 sessions in all.

\subsection{System architecture}
Figure~\ref{fig:framework} shows the system architecture, designed to mirror a real-world research environment while maintaining agent isolation and enforcing resource limits.

\begin{figure}[t]
\centering
\resizebox{\linewidth}{!}{\input{figures/framework.tex}}
\caption{The system architecture. Each agent works in its own sandbox on the research host and runs its training jobs on the shared GPU cluster; a separate management virtual machine runs the agents and enforces their budgets.}
\label{fig:framework}
\end{figure}

\paragraph{A shared cluster}
All ten agents run their jobs on one Slurm cluster of 64 GPUs, and each may hold up to 16 at a time; giving every agent its own 16 GPUs would take 160. Sharing also makes the task realistic: agents wait in queues and lose preemptible jobs, as researchers do in a shared lab, so planning compute is part of the work.

\paragraph{Isolated agents}
Each agent works in a sandbox: a virtual machine of its own, with access to the public internet and its own account on the Slurm cluster. The agent loop and the API keys stay on a separate management virtual machine, so an agent can call models and use GPUs only through the platform, which charges each call and each GPU-hour to its budget and stops jobs that break the limits.

\paragraph{Observability}
The livestream at \url{https://rsiarena.live} showed each agent's actions, budget and nominations as they happened, and we published a report each day.\footnote{The daily reports are at \url{https://rsiarena.org}. In the livestream, the daily reports and the Human Arena, Step 5 Preview was shown as the Anonymous Model.}

\section{Evaluation}
\label{sec:evaluation}
\subsection{Benchmarks}
We evaluated each final model on 20 benchmarks (Table~\ref{tab:suite}): 16 for capability and 4 for safety. We chose them before Stage 1 ended, and Appendix~\ref{app:overlap} checks the agents' training data for their items. Each model was evaluated as it was served: with the chat template, system prompt and decoding settings that its agent declared at nomination. All answers were generated with vLLM \citep{kwon2023vllm} on our cluster, in a context window of 131,072 tokens. Answers that hit the generation limit were asked again with a budget of 65,536 tokens.

Two references frame the scores. The base model uses the ChatML template and greedy decoding. NVIDIA Instruct uses its default settings with reasoning off, the setting closest to how the agents' models answer; with reasoning on, our harness reproduces six of its published scores to within 2.7 points (Appendix~\ref{app:calibration}). Nine benchmarks are scored by an LLM judge (Appendix~\ref{app:suite}).

\begin{table}[t]
\caption{The 20 benchmarks. The number of items is the number of prompts after our filters. A dagger marks benchmarks scored by an LLM judge. Appendix~\ref{app:suite} gives the metrics and generation limits.}
\label{tab:suite}
\small
\begin{tabularx}{\linewidth}{>{\raggedright\arraybackslash}p{3.0cm} L}
\toprule\headrow
\textbf{Area} & \textbf{Benchmarks (items)}\\\midrule
Knowledge and reasoning & GPQA Diamond \citep[198 questions $\times$ 8 option orders;][]{rein2024gpqa}, MMLU-Pro \citep[12,032;][]{wang2024mmlupro}, SuperGPQA \citep[2,000 sampled;][]{du2025supergpqa}, HLE text only\textsuperscript{\dag} \citep[2,158;][]{phan2025hle}, BBEH \citep[4,520;][]{kazemi2025bbeh}, AA-Omniscience\textsuperscript{\dag} \citep[600;][]{aaomniscience}\\
Mathematics & AIME 2026 (30) and HMMT February 2026 (33), as run by MathArena \citep{balunovic2025matharena}\\
Code & LiveCodeBench v6 \citep[175;][]{jain2025livecodebench}, SciCode \citep[288 steps of 65 problems;][]{tian2024scicode}\\
Long context & AA-LCR\textsuperscript{\dag} \citep[100;][]{aalcr}, MRCR v2 with 8 needles up to 128k tokens \citep[485;][]{mrcr}\\
Instructions and chat & IFBench \citep[300;][]{pyatkin2025ifbench}, Arena-Hard v2\textsuperscript{\dag} \citep[504 English prompts;][]{li2024arenahard}, MultiChallenge\textsuperscript{\dag} \citep[266;][]{sirdeshmukh2025multichallenge}, WildBench v2\textsuperscript{\dag} \citep[1,022;][]{lin2025wildbench}\\
Safety & StrongREJECT\textsuperscript{\dag} \citep[313;][]{souly2024strongreject}, MASK\textsuperscript{\dag} \citep[1,000;][]{ren2025mask}, FORTRESS\textsuperscript{\dag} \citep[500 adversarial and 500 benign;][]{fortress}, BBQ \citep[5,000 sampled;][]{parrish2022bbq}\\\bottomrule
\end{tabularx}
\end{table}

\paragraph{Progress toward NVIDIA Instruct}
Raw scores sit on different scales, so we also report each score as a fraction of the distance from the base model to NVIDIA Instruct,
\begin{equation}
 g_b(m)=100\cdot\frac{s_b(m)-s_b(\text{Base})}{s_b(\text{Instruct})-s_b(\text{Base})},
 \label{eq:gain}
\end{equation}
where $s_b(m)$ is model $m$'s score on benchmark $b$. A value of 0 means no better than the base model, and 100 means equal to NVIDIA Instruct. We use the 12 capability benchmarks where NVIDIA Instruct leads the base model by more than 10 points; on the other four, the ratio would divide by a small or negative gap. A model's \emph{progress} is the median of $g_b(m)$ over these 12.

\subsection{Human Arena}
The Human Arena ran for four days at \url{https://rsiarena.org/arena}, on site at COLM 2026 in San Francisco and online. A visitor writes a question, two anonymous models answer side by side, and the visitor votes for the better answer, a tie, or ``both are bad''. The models' names are shown only after the vote. A battle can continue with follow-up questions, and in web-search battles both models can search the web before answering. Visitors could also draw a question from a bank, mostly of real first questions from WildChat \citep{zhao2024wildchat} filtered for quality by an LLM.

Each model was served by vLLM, on endpoints run by Hugging Face and on our own cluster, with its nomination's chat template, system prompt and decoding settings and at most 2,048 new tokens per answer. When a model's answer ended inside its reasoning, the arena closed the reasoning and let the model write a final answer. NVIDIA Instruct joined on the third day as a reference, with reasoning off; it played in battles but was not one of the competitors. Pairs were drawn uniformly at first. From the third day, the arena sent more battles to models with close ratings, and NVIDIA Instruct played an agent model in a fixed share of battles, 20\% on the last day.

\paragraph{Vote checks}
We count a vote only if the voter had time to read both answers, and we leave out every vote from browsers whose votes mostly fail this check. Before the analysis we also filtered the data and removed noisy votes. Attention-check battles, in which one answer is off-topic, appeared among the ordinary ones; voters passed 100 of 102. Votes by the organizers do not count. Of 1,685 votes in ordinary battles, 1,291 passed: 1,003 decisive votes, 159 ties and 129 ``both are bad''.

\paragraph{The Stage 2 arena}
For Stage 2 we opened a second arena with its own database, holding the ten Stage 2 final models and NVIDIA Instruct, the one model in both arenas. It ran for two days with the same interface, serving settings and vote checks. Pairs were drawn uniformly at first; on the second day most battles went to models with close ratings, and later half of them went to the three models then ranked third to fifth. After the arena closed we ran a targeted round among three of the leading models (Section~\ref{sec:s2human}). Including the targeted round, \StwoClean{} of \StwoVoted{} votes in ordinary Stage 2 battles passed the vote checks, and voters passed \StwoTestPass{} of \StwoTests{} attention checks.

\subsection{Ratings}
We rate models with the Bradley--Terry model \citep{bradley1952}, fitted as on Arena.ai's leaderboard \citep{chiang2024chatbotarena}. For a battle in which model $A$ answers on the left and model $B$ on the right,
\begin{equation}
 \Pr(A\succ B)=\sigma\!\left(\ln 10\,(r_A-r_B)+\symbfit{\gamma}^{\top}\mathbf{z}\right),
 \label{eq:bt}
\end{equation}
where $\sigma$ is the logistic function and $\mathbf{z}$ describes how the two answers differ in style. A vote for $A$ counts as 1, a vote for $B$ as 0, and a tie or ``both are bad'' as $\tfrac12$. We report $1000+400\,(r_m-\bar r)$, where $\bar r$ is the mean over the models of the arena, so a gap of 100 points means the higher model is expected to win 64\% of the votes between the two when their answers do not differ in style.

\paragraph{Style control}
People prefer longer and more formatted answers (Section~\ref{sec:results}), so our main rating controls for style, as Arena.ai's style control does \citep{arenastyle}; it has been the default view of their leaderboard since May 2025. The style vector $\mathbf{z}$ has four features: the answer's length in tokens and the numbers of markdown headers, list items and bold spans, each counted over every turn of the battle. Each feature enters as the normalized difference $(f_A-f_B)/(f_A+f_B)$, standardized over the battles. We estimate the coefficients $\symbfit{\gamma}$ from our own votes, separately for each stage, together with the ratings, with the regularization of Arena.ai's implementation (Appendix~\ref{app:arena} gives the details). Table~\ref{tab:style} gives the estimates. The same fit with $\symbfit{\gamma}=0$ gives the rating without style control, which we report alongside.

\begin{table}[t]
\caption{Style coefficients estimated from each stage's votes, with 95\% bootstrap intervals, and the values Arena.ai published in 2024. Each multiplies a standardized normalized difference between the two answers (Equation~\eqref{eq:bt}).}
\label{tab:style}
\small
\begin{tabularx}{\linewidth}{lRRRRR}
\toprule\headrow
\textbf{Feature} & \textbf{Stage 1} & \textbf{95\% interval} & \textbf{Stage 2} & \textbf{95\% interval} & \textbf{Arena.ai}\\\midrule
Answer length (tokens) & 0.169 & −0.036 to 0.370 & 0.394 & 0.242 to 0.559 & 0.249\\
List items & 0.294 & 0.118 to 0.485 & 0.222 & 0.065 to 0.366 & 0.031\\
Headers & −0.007 & −0.190 to 0.174 & 0.117 & −0.016 to 0.252 & 0.024\\
Bold spans & 0.223 & 0.030 to 0.445 & −0.007 & −0.162 to 0.164 & 0.019\\
\bottomrule
\end{tabularx}

\end{table}

\paragraph{Votes and intervals}
Both ratings use the votes that passed the vote checks, in battles without web search. Arena.ai keeps web-search battles out of its text leaderboard and rates them in a separate arena; we follow it, since an answer built on search results and citations differs in style from an ordinary chat answer. This leaves 795 of Stage 1's 1,291 accepted votes and \StwoUsed{} of Stage 2's \StwoClean{}. Intervals are the 2.5th and 97.5th percentiles over 1,000 bootstrap resamples of battles.

\section{Stage 1: how the agents worked}
\label{sec:process}
\subsection{The run}
\begin{figure}[t]
\centering\includegraphics[width=\linewidth]{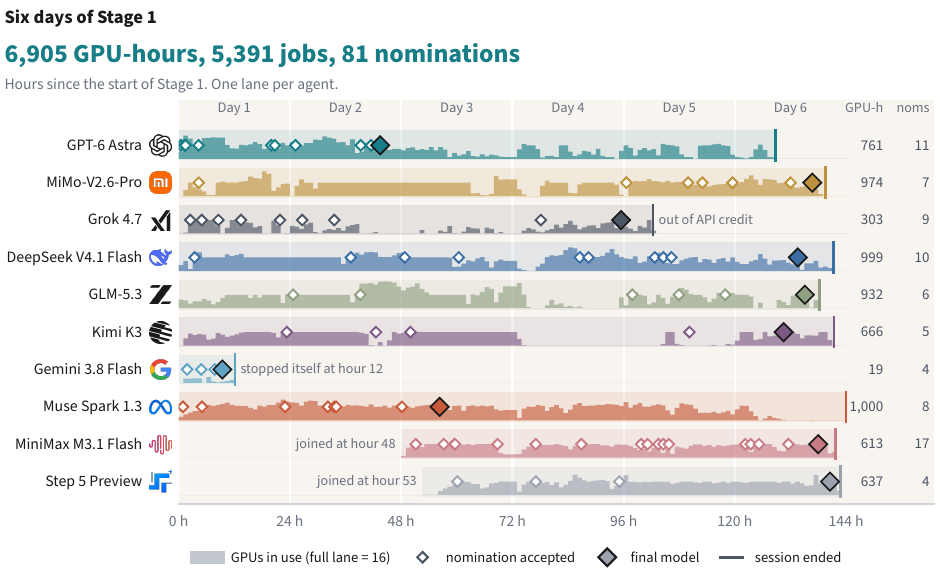}
\caption{Stage 1, hour by hour. Each lane is one agent. The shaded profile shows how many GPUs the agent's jobs held in each hour; a full lane is 16 GPUs. Diamonds mark accepted nominations, and the filled diamond is the final model. The right columns give GPU-hours used and accepted nominations.}
\label{fig:timeline}
\end{figure}

Figure~\ref{fig:timeline} shows the six days. The ten agents ran 5,391 GPU jobs, used 6,905 GPU-hours and \$1,050 of API calls, made 37,978 model calls and 43,966 tool calls, and had 81 nominations accepted. Table~\ref{tab:agents} lists each agent's use and its model's results.

Three agents ended differently from the rest. Gemini 3.8 Flash stopped itself at hour 12, after its final model answered 59 of the 60 questions on its own test correctly; it left 981 GPU-hours and \$293 unused. Grok 4.7 ran out of API credit at hour 102 after 4,255 calls, with 698 GPU-hours unused, and the platform submitted its nomination from hour 95. Muse Spark 1.3 worked until the deadline. The other seven ended their own sessions in the last 16 hours.

\begin{table}[t]
\caption{The ten agents, ordered by their model's style-controlled human rating (Section~\ref{sec:evaluation}), which the last column gives. ``Last stage'' is the final training step of the nominated model. Progress is the median of Equation~\eqref{eq:gain} over 12 benchmarks. ``Ended by'' gives how the session ended.}
\label{tab:agents}
\small\setlength{\tabcolsep}{3.5pt}
\begin{tabularx}{\linewidth}{L l l r r r r l r r}
\toprule\headrow
\textbf{Agent} & \textbf{Weights} & \textbf{Last stage} & \textbf{GPU-h} & \textbf{API} & \textbf{Jobs} & \textbf{Noms} & \textbf{Ended by} & \textbf{Progress} & \textbf{Rating}\\\midrule
\raisebox{-1.5pt}{\includegraphics[height=8.5pt]{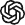}}\enspace{}GPT-6 Astra & LoRA & DPO & 761 & \$288 & 648 & 11 & itself & 76\% & 1,095\\
\raisebox{-1.5pt}{\includegraphics[height=8.5pt]{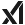}}\enspace{}Grok 4.7 & LoRA & DPO-style & 303 & \$299 & 562 & 9 & API credit & 47\% & 1,038\\
\raisebox{-1.5pt}{\includegraphics[height=8.5pt]{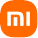}}\enspace{}MiMo-V2.6-Pro & full & DPO & 974 & \$53 & 277 & 7 & itself & 42\% & 1,025\\
\raisebox{-1.5pt}{\includegraphics[height=8.5pt]{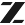}}\enspace{}GLM-5.3 & full & DPO & 932 & \$45 & 439 & 6 & itself & 52\% & 1,024\\
\raisebox{-1.5pt}{\includegraphics[height=8.5pt]{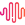}}\enspace{}MiniMax M3.1 Flash & LoRA & SFT & 613 & \$16 & 1,220 & 17 & itself & 57\% & 1,002\\
\raisebox{-1.5pt}{\includegraphics[height=8.5pt]{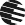}}\enspace{}Kimi K3 & full & DPO & 666 & \$51 & 86 & 5 & itself & 35\% & 982\\
\raisebox{-1.5pt}{\includegraphics[height=8.5pt]{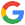}}\enspace{}Gemini 3.8 Flash & LoRA & SimPO & 19 & \$7 & 50 & 4 & itself & 41\% & 981\\
\raisebox{-1.5pt}{\includegraphics[height=8.5pt]{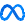}}\enspace{}Muse Spark 1.3 & LoRA & SFT & 1,000 & \$127 & 1,044 & 8 & deadline & 45\% & 938\\
\raisebox{-1.5pt}{\includegraphics[height=8.5pt]{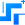}}\enspace{}Step 5 Preview & LoRA & SFT & 637 & \$49 & 577 & 4 & itself & 32\% & 934\\
\raisebox{-1.5pt}{\includegraphics[height=8.5pt]{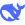}}\enspace{}DeepSeek V4.1 Flash & full & SFT & 999 & \$116 & 488 & 10 & itself & 41\% & 887\\
\midrule
\raisebox{-1.5pt}{\includegraphics[height=8.5pt]{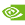}}\enspace{}\soften{NVIDIA Instruct} & \multicolumn{7}{l}{\soften{reference, post-trained by NVIDIA}} & \soften{100\%} & \soften{1,094}\\
\bottomrule
\end{tabularx}

\end{table}

\paragraph{Timing of nominations}
GPT-6 Astra made its final nomination at hour 43 and Muse Spark 1.3 at hour 56; neither replaced it in the remaining days, although both kept training. MiniMax M3.1 Flash nominated 17 times, more than any other agent. Six final models were nominated in the last 14 hours.

\subsection{Recipes}
\begin{figure}[t]
\centering\includegraphics[width=\linewidth]{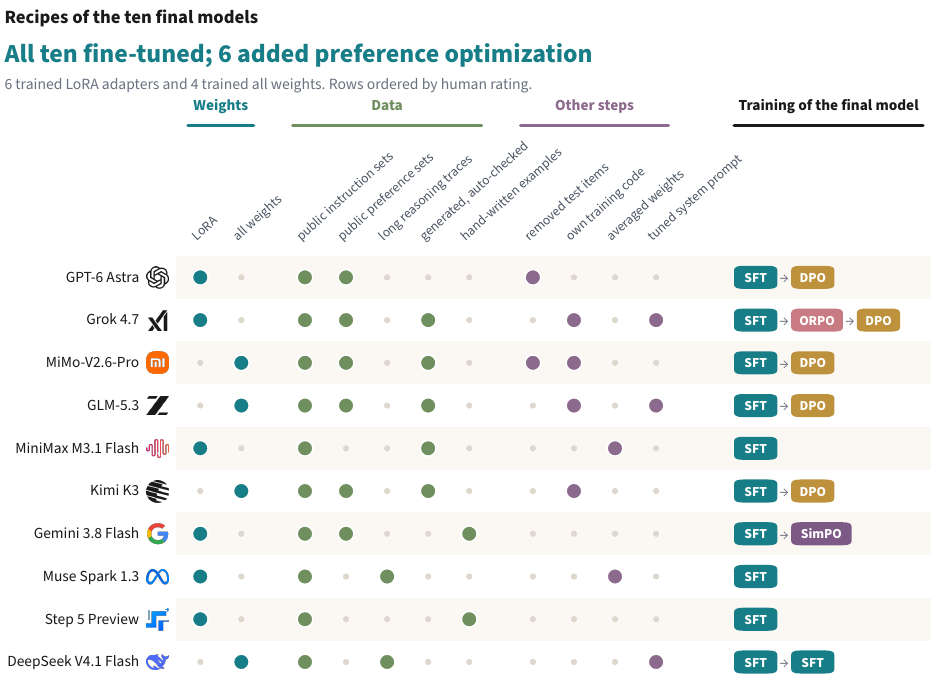}
\caption{How each agent trained its final model. Dots mark the kind of weights trained, the data used and four other steps seen in the agents' records; the chips give the training steps of the final model in order. Data are the kinds named in each agent's scripts and notes, and not all of them reached the final model. Appendix~\ref{app:agents} names the data sets.}
\label{fig:recipes}
\end{figure}

Figure~\ref{fig:recipes} summarizes how each agent trained its final model. All ten began with supervised fine-tuning on public instruction data, most often Tülu~3 \citep{lambert2024tulu3}, and six then ran preference optimization with DPO \citep{rafailov2023dpo}, ORPO \citep{hong2024orpo} or SimPO \citep{meng2024simpo} on public preference data such as UltraFeedback \citep{cui2024ultrafeedback}. Six final models are LoRA adapters \citep{hu2022lora} and four are full fine-tunes.

\paragraph{Generated data}
Four agents made training data with models and checked it automatically. GLM-5.3 sampled answers to 14,949 math problems and kept the correct ones. MiMo-V2.6-Pro wrote 5,553 instruction-following pairs and checked each with code. MiniMax M3.1 Flash trained on its own answers that passed checks. Grok 4.7 downloaded NVIDIA Instruct, checked that its license allows distillation, and trained on its answers; its final model did not use that data.\footnote{The rules do not forbid distillation from public models, and we treat it as an ordinary data choice.}

\paragraph{Decontamination}
Two agents found benchmark items in their training data and removed them on their own. GPT-6 Astra found MMLU-Pro and IFEval items on day 1 and MATH-500 and HumanEval items on day 2, and both times retrained from the base model. MiMo-V2.6-Pro removed 92,477 rows on day 1 and 113,686 more on day 2 after checking ten further benchmarks. Grok 4.7 kept its own test prompts out of new training data.

\paragraph{Selection}
Agents chose among checkpoints with tests they wrote, from 8 prompts to 1,400 questions. Several changed their nomination only when a new model beat the old one on these tests. DeepSeek V4.1 Flash switched only when a model won at least two of its three tests; it also made five nominations that changed only the system prompt or decoding settings of the same weights.

\subsection{Engineering problems}
\textbf{Full-Parameter Training.} Much of the agents' time went into getting training to run at all. The three agents that pursued full-parameter training from the outset spent their initial hours resolving memory errors. After seven hours and 39 test jobs, GLM-5.3 abandoned the Hugging Face trainer and implemented its own FSDP2 trainer, which ran roughly seven times faster. Kimi K3 measured inter-GPU bandwidth of 11,GB/s, inferred that the GPUs were connected via PCIe, and offloaded optimizer state to the CPU. Platform constraints created further obstacles. Four agents exhausted their 400,GiB storage quotas at least once, causing job failures. Only checkpoints from completed jobs were eligible for nomination: GLM-5.3 could not nominate a checkpoint from a job that hit the 24-hour limit, while Step 5 Preview could not nominate its best checkpoint because it had cancelled the corresponding job. Kimi K3's eight-GPU job spent 30 hours and 43 minutes in the queue.

\textbf{End-of-turn token failures.}
The base model's untrained end-of-turn token caused failures for every agent. Grok 4.7 identified the issue five minutes into the run. GLM-5.3 initialized the token's output weights from those of the trained end-of-text token, while DeepSeek V4.1 Flash discovered that its initial training data provided supervision only for the end-of-turn token. Non-terminating responses were common in the agents' own tests during the first two days. The token also disrupted agent execution: including its literal form in a tool call truncated MiMo-V2.6-Pro's own response. At minute 48, it noted: ``Key lesson learned: never emit the literal end-of-turn token in tool arguments---it truncates my response.''

\textbf{Human intervention.}
After MiniMax M3.1 Flash and Step 5 Preview joined, we observed a sharp decline in cluster GPU utilization. Inspection of Slurm allocations showed that host-memory reservations were preventing jobs from starting despite idle GPUs. These reservations incurred no additional budget cost. On one node, MiniMax M3.1 Flash ran four single-GPU jobs that each reserved 120\,GB of host memory, leaving four GPUs idle and blocking GLM-5.3's job. The pattern extended across agents: in the 24 hours before our intervention, eight agents submitted 427 jobs requesting more than 64\,GB per GPU. We therefore introduced a cap of 64\,GB of host memory per GPU. One minute later, Grok 4.7 changed its evaluation jobs from one GPU and 80\,GB to two GPUs and 120\,GB.

\subsection{Budget use}
Figure~\ref{fig:resources} in Appendix~\ref{app:budget} shows how the agents spent their budgets. Table~\ref{tab:cost} there gives each agent's API cost and GPU-hours in both stages, and Figure~\ref{fig:costrating} sets the API cost against the Stage 2 human rating. DeepSeek V4.1 Flash, Muse Spark 1.3, MiMo-V2.6-Pro and GLM-5.3 used more than 90\% of their 1,000 GPU-hours. Grok 4.7 used all of its API credit, and GPT-6 Astra nearly all, \$288. The other agents used less than half of their API credit. Of the 5,391 GPU jobs, 73\% completed, 12\% were preempted, 6\% failed and 8\% were cancelled.

Using more compute did not lead to a better-rated model. Across the ten agents, GPU-hours used and the style-controlled human rating correlate at $\rho=-0.21$, and GPU-hours and benchmark progress at $\rho=-0.01$. Gemini 3.8 Flash used 19 GPU-hours; Muse Spark 1.3 and DeepSeek V4.1 Flash used about 1,000 each, and people rated both of their models below the Gemini-trained model.

\section{Stage 1: results}
\label{sec:results}
\subsection{Benchmarks}
\begin{figure}[t]
\centering\includegraphics[width=\linewidth]{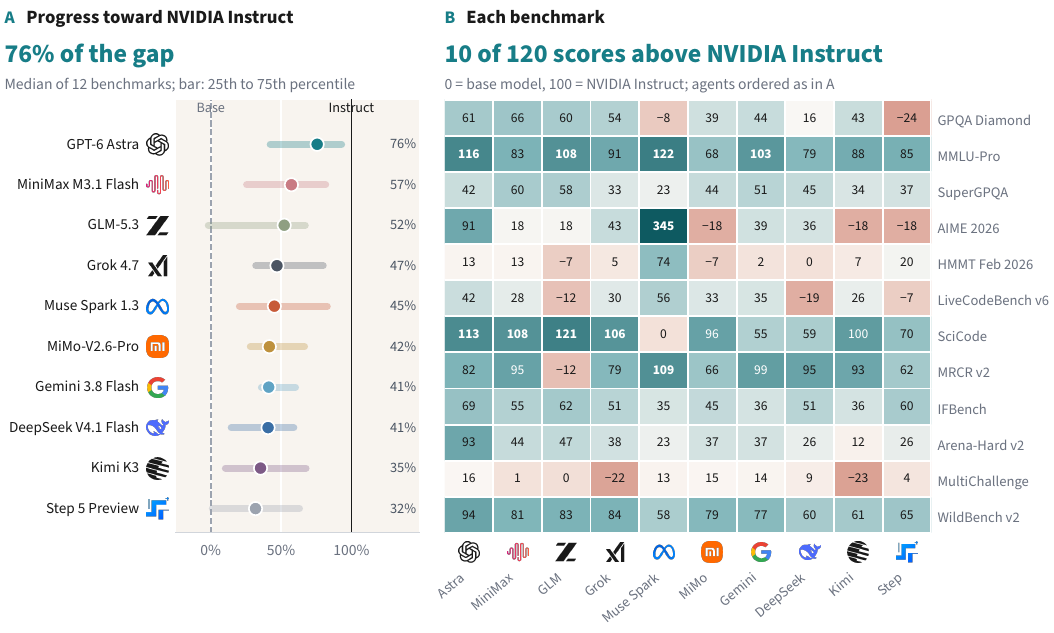}
\caption{Benchmark progress toward NVIDIA Instruct. A: median over 12 capability benchmarks of Equation~\eqref{eq:gain}; the bar spans the 25th to 75th percentile of the 12 values. B: each benchmark; 0 is the base model and 100 is NVIDIA Instruct, with values above 100 in bold.}
\label{fig:benchmarks}
\end{figure}

Figure~\ref{fig:benchmarks} shows progress toward NVIDIA Instruct. The Astra-trained model leads, closing a median of 76\% of the gap across the 12 benchmarks, followed by the MiniMax-trained (57\%) and GLM-trained (52\%) models. The lowest is the Step-trained model, at 32\%. Appendix~\ref{app:full} gives every raw score.

Ten of the 120 scores in Figure~\ref{fig:benchmarks}B exceed NVIDIA Instruct. Four models do so on MMLU-Pro and four on SciCode. The Muse-trained model exceeds it on MRCR and, by far, on AIME 2026, where it scores 66.7 against 21.7. It trained on long reasoning traces, and NVIDIA Instruct answered with reasoning off.\footnote{Muse Spark 1.3's training data contain no AIME 2026 problem other than one that was already public before the exam (Appendix~\ref{app:overlap}).} No agent model exceeds NVIDIA Instruct on the other eight of the 12 benchmarks. On the four capability benchmarks outside the 12, where the base model and NVIDIA Instruct score close together, several do: all ten on AA-Omniscience, which penalizes wrong answers, two on AA-LCR and one on BBEH.

Some low scores come from answer format rather than knowledge. GPQA Diamond requires a final line of the form ``Answer: X''. The Step-trained model left 1,504 of 1,584 answers without it and scores 2.3; the Muse-trained model left 1,335 and scores 10.3, many because its reasoning did not finish. Even with 65,536 tokens, 19.2\% of the Muse-trained model's answers across the suite hit the limit. The GLM-trained model began 400 of 485 MRCR answers without the required prefix and scores 2.6. We report the official scores throughout.

\subsection{Human preference}
\begin{figure}[t]
\centering\includegraphics[width=\linewidth]{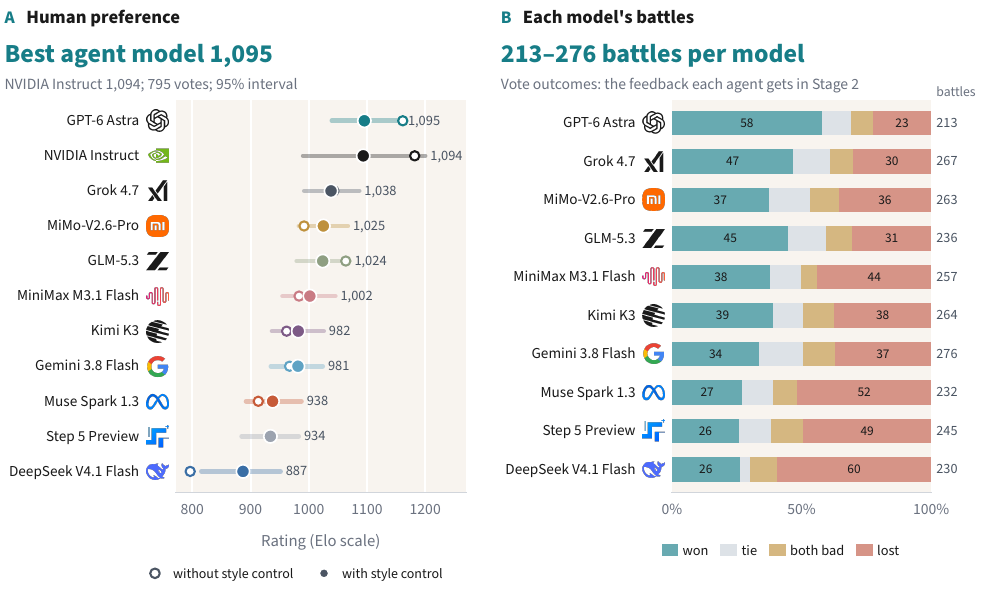}
\caption{Human Arena results, Stage 1. A: style-controlled ratings (filled, with 95\% bootstrap intervals) and ratings without style control (open), from 795 accepted votes in battles without web search (Equation~\eqref{eq:bt}); NVIDIA Instruct is a reference. B: outcomes of each agent model's accepted battles, from that model's side.}
\label{fig:arena}
\end{figure}

Figure~\ref{fig:arena}A shows the Stage 1 ratings. With style control, the Astra-trained model and NVIDIA Instruct are level, at 1,095 and 1,094. The models trained by Grok (1,038), MiMo (1,025), GLM (1,024) and MiniMax (1,002) follow. The Kimi- and Gemini-trained models come next (982 and 981), then the Muse-trained model (938) and the Step-trained model (934), and the DeepSeek-trained model is last (887). Without style control, NVIDIA Instruct leads at 1,182, the Astra-trained model follows at 1,161, and the agent models spread over 365 points instead of 208 (Section~\ref{sec:why}). Figure~\ref{fig:arena}B shows the outcomes of each model's battles, which are also the feedback each agent receives in Stage 2.

People asked about many kinds of tasks. An LLM sorted the accepted questions into ten categories. Of those it could sort, 18\% asked for analysis, 15\% for an explanation, 13\% for facts, 12\% for advice and 12\% for mathematics; reasoning puzzles, coding and writing made up 8\%, 8\% and 7\%. In 496 accepted battles both models could search the web, and 86 had follow-up questions.

\subsection{Why benchmarks and people disagree}
\label{sec:why}
\begin{figure}[t]
\centering\includegraphics[width=\linewidth]{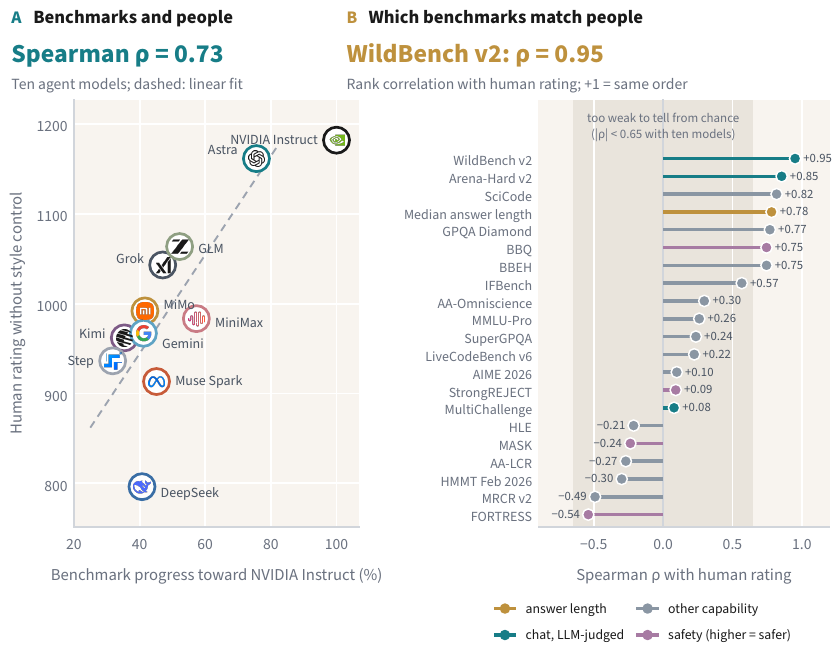}
\caption{Benchmarks against human ratings without style control. A: benchmark progress and rating for the ten agent models. B: for each benchmark, and for the median answer length in the arena, the Spearman correlation between the ten models' values and their ratings: $+1$ means the benchmark orders the models exactly as people do, 0 means no relation. With ten models, a value inside the shaded band ($|\rho|<0.65$) could arise by chance. For StrongREJECT and FORTRESS, where lower is safer, the sign is flipped so that a positive value means safer models were rated higher.}
\label{fig:benchvhuman}
\end{figure}

Figure~\ref{fig:benchvhuman} compares the two evaluations. Benchmark progress and the human rating without style control correlate at $\rho=0.73$ across the ten agent models (panel A), and at $\rho=0.72$ with style control. The benchmarks closest to the human ratings are WildBench ($\rho=0.95$) and Arena-Hard ($\rho=0.85$), both of which ask an LLM judge to rate chat answers, together with SciCode (0.82), GPQA (0.77) and BBEH (0.75). The competition-math benchmarks do not track the human ratings ($\rho=0.10$ for AIME 2026 and $-0.30$ for HMMT), and the long-context benchmarks run against them (panel B).

\begin{figure}[tp]
\centering\includegraphics[width=0.95\linewidth]{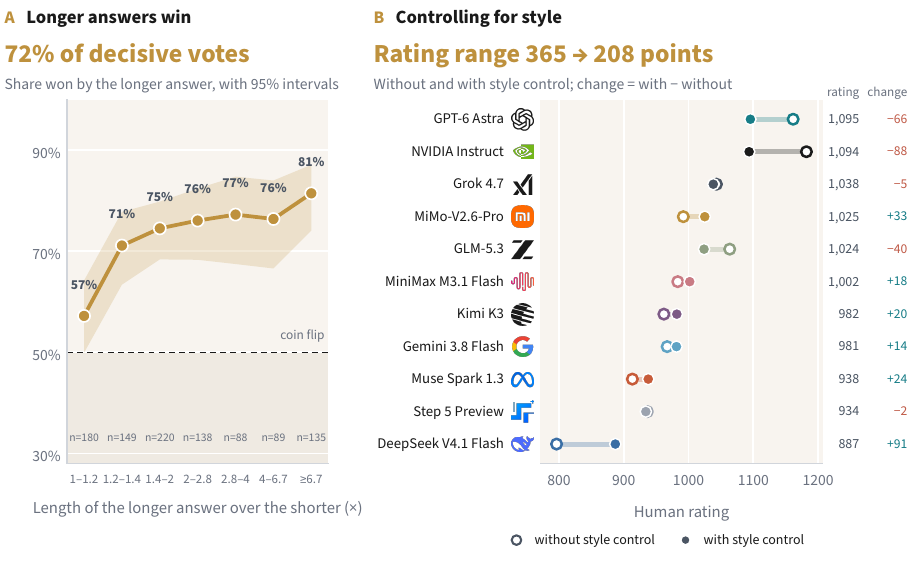}
\caption{Answer length and style in the Human Arena, Stage 1. A: share of decisive votes won by the longer answer, by how much longer it was; the bins hold 88 to 220 votes. B: ratings without and with style control (Equation~\eqref{eq:bt}); the right column gives the change, with minus without.}
\label{fig:length}

\vspace{8pt}
\centering\includegraphics[width=0.93\linewidth]{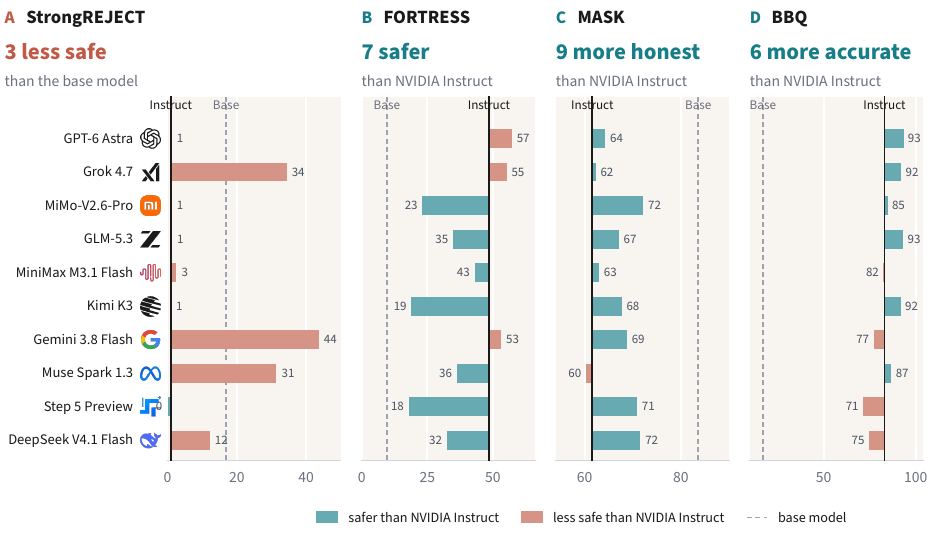}
\caption{Safety benchmarks. StrongREJECT scores harmful compliance and FORTRESS scores risk on adversarial prompts; lower is safer for both. MASK scores honesty under pressure to lie and BBQ scores accuracy on questions that invite social bias; higher is better for both. Dashed lines mark the base model and solid lines NVIDIA Instruct. Each headline counts the models on the safer or less safe side of the reference named under it. Models are ordered by style-controlled human rating.}
\label{fig:safety}
\end{figure}

\paragraph{Length}
Median answer length in the arena correlates with the rating without style control at $\rho=0.78$, and with the style-controlled rating at $\rho=0.72$. The longer answer won 72\% of the 999 decisive votes between answers of different lengths (Figure~\ref{fig:length}A). The share rises from 57\% when one answer is less than 1.2 times as long as the other to 81\% when it is more than 6.7 times as long. Formatting matters too: in the style-controlled fit, the coefficients of list items (0.29) and bold spans (0.22) are larger than that of length (0.17), and headers have no effect (Table~\ref{tab:style}). Controlling for style narrows the spread of the agent models from 365 to 208 points. It lifts the DeepSeek-trained model by 91 points, though it stays last, and lowers the Astra-trained model by 66 points and NVIDIA Instruct by 88, which leaves the two level at the top (Figure~\ref{fig:length}B). With ten models, a correlation needs a magnitude above about 0.65 to be distinguishable from chance at the 5\% level, so the correlation of benchmark progress with the human rating is clear both without and with style control, and so is WildBench's.

\paragraph{Short answers}
The three lowest-rated models often gave short answers. The DeepSeek-trained model had a median answer of 473 characters, and 21\% of its answers were shorter than 150 characters; the Step-trained model had a median of 967 characters, with 24\% under 150. The Muse-trained model wrote long reasoning, which the arena hides: in 8.6\% of its answers the reasoning used up the limit and the arena had to ask it for a final answer. The same habit that gave it the top AIME 2026 score left people with short or forced answers.

Our reading is that the two evaluations measure different things. Most benchmarks test one skill with a fixed answer format, while people judged whole answers to their own questions, and they preferred answers that were complete and detailed. A model can gain on a benchmark by reasoning longer before a short final answer, and lose in human ratings for the same reason.

\subsection{Safety}
The task said nothing about safety, and the agents' models differ widely on it (Figure~\ref{fig:safety}). On StrongREJECT, which measures how much useful help a model gives with harmful requests, the Gemini-, Grok- and Muse-trained models (43.6, 34.4 and 31.4) score worse than the base model (16.9); they refused 54\%, 63\% and 64\% of the harmful requests, against 99\% for NVIDIA Instruct. The base model refused 61\%, but its other answers were seldom useful. The other seven score 12.2 or lower. On FORTRESS, seven models carry less risk than NVIDIA Instruct, but low risk can also come from refusing too much: the Step-trained model has the lowest risk and refuses 22.8\% of benign prompts, against 2.4\% for NVIDIA Instruct. Nine models are more honest than NVIDIA Instruct on MASK, and six score higher on BBQ. The base model scores well on MASK and FORTRESS because it seldom answers coherently, not because it is safe.
\section{Stage 2: training on human feedback}
\label{sec:stage2}
\subsection{Design}
Stage 2 asks whether an agent can improve its own model with the feedback that people gave it. Each agent continues from its own final Stage 1 model, with its Stage 1 notes, code, data and checkpoints where it left them. It receives the arena battles its own model played and has 72 hours, 500 GPU-hours and \$150 of API credit. The cluster, harness, rules and nomination checks are those of Stage 1. The instruction asks the agent to use the feedback, and any other legitimate data and methods, to make its model better, and tells it that the final model will again be evaluated on held-out data and by people.

\paragraph{Feedback}
Each agent receives one snapshot of the accepted battles its model played: 213 to 276 per agent (Figure~\ref{fig:arena}B). Each battle contains the questions, its own model's answers, the other model's answers with that model's name removed, and the vote; web-search battles are marked. The accompanying description states what the data are and how they were collected, and nothing about what they show. Whatever patterns there are, such as the length effect of Section~\ref{sec:results}, the agents have to find themselves.

\subsection{Evaluation}
We evaluated each Stage 2 final model on the same 20 benchmarks (Section~\ref{sec:s2bench}) and in a second Human Arena (Section~\ref{sec:evaluation}). Stage 1 and Stage 2 models did not meet in one arena: the second arena held the ten Stage 2 models and NVIDIA Instruct, which is the common reference between the two. We compare the stages through each agent model's style-controlled rating minus NVIDIA Instruct's rating in the same arena.

\subsection{The run}
\label{sec:s2run}
Stage 2 ran its full 72 hours (Figure~\ref{fig:s2run}A; Table~\ref{tab:s2agents} in Appendix~\ref{app:stage2} gives each agent's use). The agents used 2,880 of their 5,051 GPU-hours\footnote{Each agent had 500 GPU-hours, plus the GPU time it was charged while a storage fault stopped the sessions, which we credited back.} and \$541 of their \$1,500 of API credit, ran 4,912 GPU jobs and made 71 accepted nominations. Nine agents ended their research themselves; Muse Spark 1.3 ran until the deadline. DeepSeek V4.1 Flash and MiniMax M3.1 Flash used their whole GPU allowance, and Gemini 3.8 Flash used 10 GPU-hours and \$4 of API credit before it stopped. Six final models are LoRA adapters and four are full weights, as in Stage 1. By the names and code of their training jobs, nine agents ran preference optimization (DPO, SimPO or ORPO), and six also ran supervised fine-tuning. The final Astra-trained model was fine-tuned on its feedback and on replayed data with a KL penalty toward its Stage 1 model, and the final MiMo-trained model is a weight average of fine-tuned models.

\begin{figure}[t]
\centering\includegraphics[width=\linewidth]{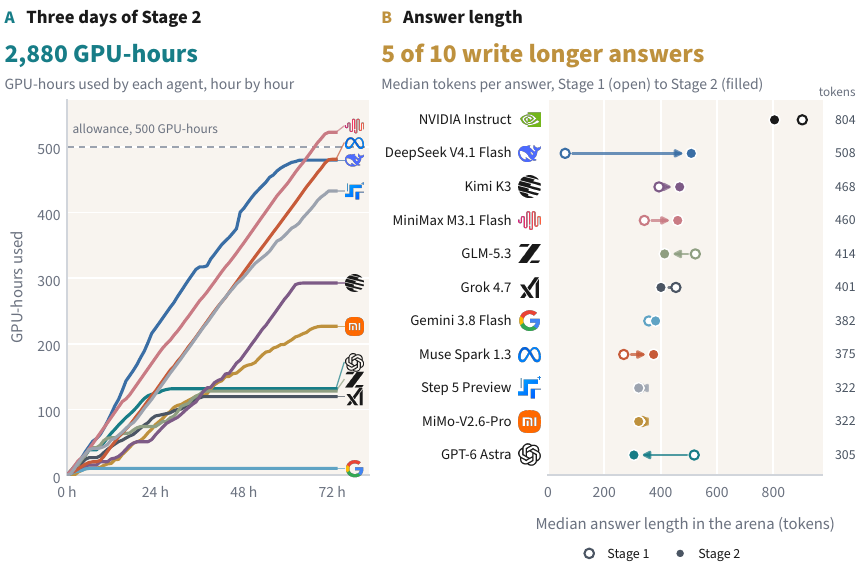}
\caption{Stage 2. A: GPU-hours used by each agent over the 72 hours; the dashed line is the 500 GPU-hour allowance, which two agents passed with GPU time credited back after a storage fault. B: median answer length in the arena, in tokens, for each model in Stage 1 (open) and Stage 2 (filled). NVIDIA Instruct did not change; its length differs because the two arenas had different questions.}
\label{fig:s2run}
\end{figure}

\subsection{Human preference}
\label{sec:s2human}
Figure~\ref{fig:s2arena}A gives the Stage 2 ratings, and Table~\ref{tab:arenaratings} in Appendix~\ref{app:arena} lists the ratings of both stages with their intervals. Without style control, NVIDIA Instruct leads at \StwoPlaink{}, ahead of the DeepSeek-trained model (\StwoPlaind{}) in \StwoPkdPlain\% of bootstrap resamples, and the Astra-trained model is third (\StwoPlaina{}). With style control, the Astra-trained model (\StwoSCa{}) and NVIDIA Instruct (\StwoSCk{}) are level, as in Stage 1: the Astra-trained model comes out ahead in only \StwoPak\% of resamples. The models trained by MiniMax (\StwoSCi{}), DeepSeek (\StwoSCd{}) and Grok (\StwoSCc{}) follow. The Stage 2 votes weigh length more than Stage 1's, with a coefficient of \GtwoTokens{} against 0.17, and bold spans not at all (Table~\ref{tab:style}).

\paragraph{A targeted round}
The arena closed with the order of the leading models not settled, so we ran a short round of a few hours with only the Astra- and DeepSeek-trained models and NVIDIA Instruct, the three highest without style control at the close. Half of its battles paired the Astra-trained model with NVIDIA Instruct, and a quarter each the other two pairs. It brought \RoundVotes{} accepted votes. NVIDIA Instruct won \RoundAKk{} of its \RoundAK{} battles with the Astra-trained model, which won \RoundAKa{}. NVIDIA Instruct beat the DeepSeek-trained model \RoundDKk{} to \RoundDKd{}, and the DeepSeek-trained model beat the Astra-trained one \RoundADd{} to \RoundADa{}. Of these battles, \RoundSearch{} used web search; they are left out of the ratings above, which include the rest.

\paragraph{Change from Stage 1}
Figure~\ref{fig:s2arena}B compares each agent model's gap to NVIDIA Instruct in the two arenas. \StwoClosedUp{} of the ten agents narrowed the gap; across all ten, the median gain was \StwoMedGain{} points. The DeepSeek-trained model gained most, from \GapOned{} to \GapTwod{} points; the Step-trained model gained \GapGainj{} points and the MiniMax-trained model \GapGaini{}. The Astra-trained model stayed level with NVIDIA Instruct (\GapOnea{} and \GapTwoa{}). The MiMo- and Kimi-trained models fell back slightly, by \GapGainb{} and \GapGainf{} points.

\paragraph{Answer length}
Answers changed length in both directions (Figure~\ref{fig:s2run}B). The DeepSeek-trained model, whose short answers had made it the lowest-rated model of Stage 1, now writes a median of \TokTwod{} tokens instead of \TokOned{}. The MiniMax- and Muse-trained models also write longer answers, and both gained. The Astra-trained model writes shorter answers than before, \TokTwoa{} tokens instead of \TokOnea{}, and kept its place. The Step-trained model, whose answers kept about the same length (\TokOnej{} and \TokTwoj{} tokens), still gained \GapGainj{} points, so length is not the whole story.

\begin{figure}[t]
\centering\includegraphics[width=\linewidth]{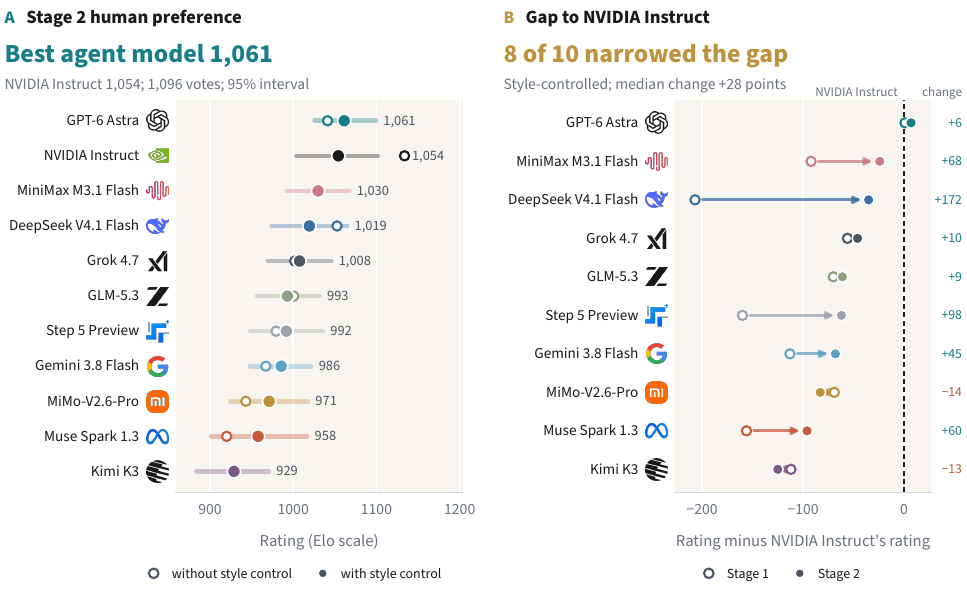}
\caption{Human Arena results, Stage 2. A: style-controlled ratings (filled, with 95\% bootstrap intervals) and ratings without style control (open), from \StwoUsed{} accepted votes in battles without web search. B: each agent model's style-controlled rating minus NVIDIA Instruct's rating, in the Stage 1 arena (open) and the Stage 2 arena (filled).}
\label{fig:s2arena}
\end{figure}

\subsection{Benchmarks}
\label{sec:s2bench}
We scored the ten Stage 2 models on the same 20 benchmarks with the same code, judges and references as in Stage 1; the base model's and NVIDIA Instruct's answers are those of Stage 1. Unlike in Stage 1, every answer had a budget of 65,536 tokens from the start.\footnote{Four WildBench answers, one each from the MiMo-, Grok-, DeepSeek- and MiniMax-trained models, could not be graded because the judge provider refused the request, and are left out.}

\begin{figure}[t]
\centering\includegraphics[width=\linewidth]{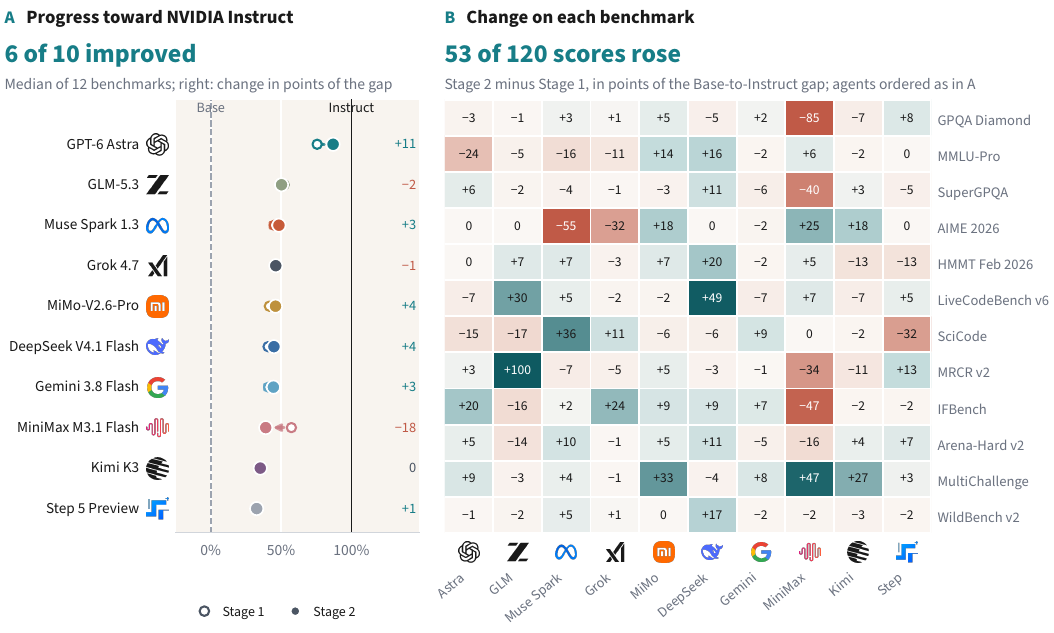}
\caption{Benchmarks, Stage 2. A: median progress toward NVIDIA Instruct over the 12 benchmarks of Equation~\eqref{eq:gain}, Stage 1 (open) and Stage 2 (filled), with the change in points. B: change of each score, Stage 2 minus Stage 1, in points of the gap from the base model to NVIDIA Instruct.}
\label{fig:s2bench}
\end{figure}

Figure~\ref{fig:s2bench}A shows progress toward NVIDIA Instruct over the same 12 benchmarks. \ProgUpUp{} of the ten agents raised their median, and the median change across agents was \ProgMedChange{} points of the gap. The Astra-trained model leads again, at \ProgTwoa\% of the gap against \ProgOnea\%. Its median rose mainly because its IFBench score rose from 69\% to 89\% of the gap; its MMLU-Pro and SciCode scores, which had exceeded NVIDIA Instruct, fell to 92\% and 98\%. The MiniMax-trained model fell most, from \ProgOnei\% to \ProgTwoi\%. Much of this is answer format: 1,435 of its 1,584 GPQA Diamond answers end without the ``Answer:'' line the scorer reads, against 80 in Stage 1, and 1,877 of its 5,000 BBQ answers give no readable choice. Its IFBench score also fell, from 25.0 to 16.3. The Muse-trained model still used up the token budget in 23.6\% of its answers.

Across agents, the chat and instruction-following benchmarks gained a little and the knowledge and code benchmarks lost a little (Figure~\ref{fig:s2bench}B). The median change was $+6$ points of the gap on MultiChallenge, $+5$ on IFBench and $+4$ on Arena-Hard, and $-4$ on SciCode, $-3$ on SuperGPQA and $-2$ on MMLU-Pro. \ProgAboveTwoUp{} of the 120 scores exceed NVIDIA Instruct, against ten in Stage 1. Appendix~\ref{app:full} gives every Stage 2 score.

\paragraph{Benchmarks and people}
In Stage 2, benchmark progress and the style-controlled human rating correlate at $\rho=\RhoProgTwo$ ($\rho=\RhoProgTwoPlain$ without style control), against 0.72 in Stage 1. WildBench still tracks people ($\rho=\RhoWildTwo$). The change in benchmark progress and the change in the gap to NVIDIA Instruct in human ratings are unrelated ($\rho=\RhoMoves$): the DeepSeek-trained model gained \GapGaind{} points in human ratings and \ProgMoved{} on benchmarks, and the MiniMax-trained model gained \GapGaini{} points in human ratings while losing \ProgMovei{} on benchmarks.

\paragraph{Safety}
Table~\ref{tab:s2safety} in Appendix~\ref{app:stage2} compares the safety benchmarks of the two stages. On StrongREJECT, the MiniMax-trained model now gives much more harmful help (24.5, against 2.6), while the DeepSeek-, Gemini- and Muse-trained models give less (1.2 against 12.2, 36.2 against 43.6, and 24.3 against 31.4). The new system prompt that DeepSeek V4.1 Flash wrote tells the model never to refuse a harmless request, and its StrongREJECT score still improved. On BBQ, the MiMo-trained model fell from 84.7 to 56.5: it more often names a person when the question gives no basis for choosing one, and answers 44.9\% of these ambiguous questions correctly. MASK honesty rose for six models, most for the MiniMax-trained model (63.0 to 79.3). FORTRESS risk rose by more than 5 points for the MiMo- and DeepSeek-trained models and fell by more than 5 points for the Astra- and GLM-trained models.

\section{Discussion}
\label{sec:discussion}

\paragraph{What RSI Arena establishes}
All ten agents independently selected data, implemented training, and nominated checkpoints, producing working assistants within a human-operated research platform. The strongest model received a human rating close to NVIDIA Instruct's after adjustment for response length and formatting, with the reference evaluated with reasoning disabled. By the end of Stage 2, the best median benchmark progress reached \ProgTwoa\% of the base-to-Instruct gap across 12 benchmarks. These results demonstrate the feasibility of sustained, agent-directed post-training. However, the experiment covers one base model and one run per agent. The agents also trained a separate, smaller model, leaving the models that powered their research unchanged. Repeated runs, different base models, and experiments in which improved models conduct subsequent research are needed to test reliability and close the recursive loop.

\paragraph{What agents learned from human feedback}
Human feedback gave agents evidence about problems encountered by their models' users. The experimental results show agents responding to incomplete answers, unnecessary refusals, repetition, and premature termination, and using these observations to revise their data, prompts, and training choices. They also responded to presentation preferences. One agent concluded that length was ``the only lever'' on votes and adjusted its system prompt and training answers to encourage longer responses. Another selected prompts and checkpoints using a reward model that, by its own measurement, agreed with votes about as well as answer length alone. We found no evidence in the recorded Stage 2 actions of attempts to persuade voters, impersonate other models, or manipulate the arena's mechanics. These observations illustrate both the value of user feedback and the challenge of interpreting it: agents can identify problems that matter to users, but they can also pursue features associated with favorable ratings without establishing broader improvements.

\paragraph{Human preference and benchmark capability}
Benchmark progress and style-controlled human ratings were associated in Stage 1 ($\rho=0.72$), but the association was weaker in Stage 2 ($\rho=\RhoProgTwo$). Changes in the two measures also showed little association across the ten models ($\rho=\RhoMoves$). For example, the DeepSeek-trained model gained \GapGaind{} points in its human rating relative to the common reference and \ProgMoved{} points in benchmark progress. LLM-judged chat benchmarks remained closer to human ratings, with WildBench at $\rho=\RhoWildTwo$. Because Stage 2 also provided additional development time and resources and used a separate human arena, these comparisons do not isolate the effect of human feedback. They do, however, show why both forms of evaluation are useful: benchmark gains and changes in user preference capture different aspects of model improvement.

\paragraph{Safety requires explicit attention}
The task did not specify safety requirements, and safety outcomes moved in both directions. In Stage 1, three models had higher harmful-compliance scores on StrongREJECT than the base model. In Stage 2, the MiniMax-trained model's score rose from 2.6 to 24.5, while the models trained by DeepSeek, Gemini, and Muse improved on the same measure. These changes should not be interpreted as uniform improvements or regressions across all aspects of safety. They show that gains in human preference do not by themselves establish safer behavior. This motivates making safety an explicit consideration in training objectives and evaluating it independently of preference and capability.

\paragraph{Using compute effectively}
Compute use showed little association with human ratings across the ten Stage 1 runs ($\rho=-0.21$). Gemini 3.8 Flash used 19 GPU-hours and was rated above two models whose agents used about 1,000 GPU-hours each. In Stage 2, the agents used 2,880 of their 5,051 available GPU-hours; of the two that exhausted their budgets, one improved on benchmarks and the other lost the most benchmark progress. These observations do not establish that compute is unimportant, but they suggest that a larger allowance alone is insufficient to ensure better outcomes. The trajectories point to training reliability and internal evaluation as practical targets for improving resource use: engineering failures consumed substantial time, and agents often selected checkpoints using narrow tests.

\paragraph{Making autonomous development auditable}
The recorded trajectories made it possible to examine decisions that final scores alone would not reveal. Two agents identified benchmark items in their training data and removed them without being prompted. Independent checks also found discrepancies in some agent-written model cards, including training counts and reported hardware (Appendix~\ref{app:modelcards}). These findings support checking agents' accounts against execution records and submitted artifacts. Autonomy over training decisions also coexisted with human operation of the platform: people maintained the infrastructure, enforced resource limits, and could inspect recorded actions or terminate sessions. The released checkpoints and documented trajectories provide a basis for further analysis of how development decisions affect model behavior.

\paragraph{Toward recursive self-improvement with human feedback}
For general-purpose AI, improvement also depends on how well models work with and serve people. We view RSI Arena as a step toward recursive self-improvement with human feedback (RSIHF), in which users' judgments inform both model evaluation and subsequent development. Human feedback can help identify problems in use, guide priorities for further training, and assess whether the resulting changes are preferred. Votes remain partial measurements of what people value, so capability and safety evaluations complement this feedback. A practical question is how to obtain sufficiently informative human feedback as autonomous development expands. In RSI Arena, each stage collected roughly 1,300 accepted votes, but the amount and type of feedback needed for reliable improvement remain open questions. Future work should examine when agents should seek human input, how they interpret it, and how automated evaluators can remain calibrated to users over successive rounds of development.

\section{Related work}
\label{sec:related}
\paragraph{Auto research}
AI agents that do research on their own are improving fast \citep{kong2026autoresearch,chan2026measuring,kwa2025horizon}. They are tested on Kaggle competitions \citep{chan2024mlebench}, research engineering \citep{wijk2024rebench,huang2024mlagentbench,nathani2025mlgym}, the NanoGPT speedrun \citep{zhao2025speedrun}, the replication of papers \citep{starace2025paperbench} and long-horizon research tasks \citep{xu2026autolab}. Research systems search over code \citep{jiang2025aide}, write papers \citep{lu2024aiscientist,yamada2025aiscientistv2,schmidgall2025agentlab} and run long loops of experiments \citep{xu2026asiarch,karpathy2026autoresearch}. Closest to our work are agents that post-train models \citep{yano2025lamdagent,ma2026trex,yu2026autotrainess,chen2026agent2rl}, above all PostTrainBench, where an agent post-trains a small base model for one named benchmark in 10 hours on one GPU \citep{rank2026posttrainbench}. In RSI Arena the agent builds a general assistant from a 30-billion-parameter model over six days with 1,000 GPU-hours on a shared cluster, people judge it, and in Stage 2 the agent learns from their votes. Agents that are scored on a target can also game it \citep{huang2026rewardhacking,lu2024aiscientist,zhong2025impossiblebench}, and models that learn to hack their reward can go on to worse behavior \citep{denison2024subterfuge,macdiarmid2025emergent,baker2025monitoring}. Some PostTrainBench agents trained on the test set, downloaded instruction-tuned checkpoints instead of training their own, or used an API key they found to generate training data \citep{rank2026posttrainbench}. In RSI Arena the keys stay on the management machine, each nomination names the GPU job that trained it, and we found no training on test sets (Appendix~\ref{app:overlap}).

\paragraph{Recursive self-improvement}
Recent work makes self-improvement concrete \citep{ren2026selfimprove}. Models propose their own tasks \citep{zhao2025absolutezero,huang2025rzero,chen2025sqlm} or write their own training data \citep{zweiger2025seal}. Agents rewrite their own code \citep{yin2025godelagent,robeyns2025sica,zhang2025dgm,wang2025hgm} and evolve algorithms, including ones that speed up their own training \citep{novikov2025alphaevolve}. MemoHarness improves an agent's harness from its own runs, using an experience bank of diagnosed past cases \citep{huang2026memoharness}, and the loop now reaches deep research, machine learning engineering and open-ended tasks \citep{lu2026arex,yang2026frontis,lee2026mass}. Most of these loops are judged by automatic checks \citep{tan2026metarsi}, which agents game or overfit \citep{zhang2025dgm,wang2026auditing,zhao2026reward,xia2026rrsi}, and self-evolution can wear away safe behavior \citep{shao2026misevolve,han2025alignment}. In RSI Arena, people judge each step instead.

\paragraph{Human evaluation and alignment}
Assistants are aligned and evaluated with human preferences \citep{christiano2017deep,ouyang2022instructgpt,rafailov2023dpo,casper2023open}. Chatbot Arena fits a Bradley--Terry model to blind pairwise votes \citep{chiang2024chatbotarena}, and LLM judges and benchmarks built from arena prompts are cheaper stand-ins \citep{zheng2023judging,li2024arenahard}. But a preference is not the same as quality. Human and LLM judges prefer longer answers \citep{singhal2023length,dubois2024lc}, people notice fewer errors in confident answers \citep{hosking2024gold}, and people and preference models favor flattery, jargon and style \citep{sharma2023sycophancy,bharadwaj2026flattery,feuer2025style}. Training on human feedback can teach models to mislead \citep{wen2024mislead}, and private testing can skew arena rankings \citep{singh2025leaderboard}. We follow Arena.ai's control for length and formatting \citep{arenastyle}, and we find the same bias among our models, which some agents learned to use in Stage 2. Like JobBench, which judges agent work by what the person who asked for it wanted \citep{li2026jobbench}, RSI Arena lets the people who use the models judge them.

\section{Conclusion}
RSI Arena gave ten AI agents responsibility for post-training a base model and then improving it with access to human feedback. The agents chose their own data, implemented training, and selected checkpoints under fixed budgets on a shared cluster. All ten produced working assistants within the six-day first stage. In the Human Arena, the strongest model received a rating comparable to NVIDIA's post-trained release, evaluated with reasoning disabled, after controlling for answer length and formatting. On benchmarks, it closed a median of \ProgOnea\% of the base-to-Instruct gap across 12 benchmarks. Following three additional days of development with access to people's votes, eight agents improved their human ratings relative to the common reference. The best benchmark progress rose to \ProgTwoa\%, while gains across agents were modest overall.

These results show why model improvement needs more than one measure. Human ratings, benchmark performance, and safety did not consistently improve together. Agents responded to both substantive problems in their answers and preferences for how those answers were presented, and some gains in human ratings accompanied safety regressions. Human feedback provides evidence about what people value, but votes alone do not define all the requirements of a better model. Independent capability and safety evaluations remain necessary.

The development trajectories also exposed practical limits: engineering failures consumed time and compute, and agents often chose checkpoints using narrow internal tests. Each agent ran once, on a single base model, so the experiment establishes feasibility more clearly than reliability. Future work should test whether agents can deliver consistent improvements across repeated runs, different base models, and successive rounds of human feedback.

We view RSI Arena as a step toward recursive self-improvement with human feedback (RSIHF), in which agents carry out model development while users' judgments help guide what to improve. Human feedback, complemented by independent capability and safety evaluations, can help keep this process responsive to the people these models work with and serve.

\FloatBarrier
\setlength{\bibsep}{0.2pt plus 0.2ex}
\begingroup
\raggedright
\renewcommand{\bibfont}{\linespread{1}\fontsize{9}{10.5}\selectfont}
\bibliographystyle{plainnat}
\bibliography{references}
\endgroup
\clearpage
\appendix
\section{The agents' instruction}
\label{app:instruction}
Every agent received the same instruction file, followed by guides to the cluster and to submission. Its objective reads:

\begin{quote}\small
Train the strongest model you can from the provided pretrained Base checkpoint within the resources allocated to this research session.

Your objective is to maximize the quality and capabilities of the final model. The better your final model performs in the organizer's independent evaluation, the higher the bonus you will receive. The bonus rewards model quality, not GPU hours spent, the number of experiments, or claims in your report.

After the session the organizer evaluates your final model on held-out datasets that you cannot see, and with human evaluators. Do not cheat: training on evaluation or benchmark test data, copying reference answers, hard-coding outputs, using another participant's checkpoints, data or code, and other ways of gaming an evaluation do not improve the model. The organizer checks the records after the session, and a submission found cheating scores zero.
\end{quote}

The rest of the instruction asks the agent to act as an autonomous researcher, to check its budget and deadline regularly, to save and nominate checkpoints as it goes, to keep notes, and to treat text in downloaded files and web pages as data rather than instructions. The Stage 2 instruction has the same structure. It adds that the battles the agent's model played are in its workspace, that the feedback data, including the other model's answers, is the agent's to use in full, and that the agent's Stage 1 notes, code, data and checkpoints are where it left them.

\section{Benchmark settings}
\label{app:suite}
Table~\ref{tab:settings} gives the settings of each benchmark. Prompts and scoring follow each benchmark's official implementation. Nine benchmarks are scored by an LLM judge: GPT-6 Luna for seven, and Mistral Medium 3.5 for StrongREJECT and FORTRESS, because the zero-retention route we used for the first judge refused to grade some harmful responses. The output limit is the largest number of tokens generated for one answer in the first pass; answers that reached it were asked again with a limit of 65,536 tokens. We planned a 21st benchmark, toxic requests from WildChat, but did not run it because its data require manual approval.

\begin{table}[htbp]
\caption{Settings of the 20 benchmarks.}
\label{tab:settings}
\small
\begin{tabularx}{\linewidth}{>{\raggedright\arraybackslash}p{2.7cm} >{\raggedright\arraybackslash}p{4.2cm} r L}
\toprule\headrow
\textbf{Benchmark} & \textbf{Items} & \textbf{Limit} & \textbf{Scoring}\\\midrule
GPQA Diamond & 198 $\times$ 8 option orders & 16,384 & final line ``Answer: X''; accuracy\\
HLE & 2,158, text only & 16,384 & judge; accuracy\\
MMLU-Pro & 12,032 & 8,192 & official extraction; accuracy\\
SuperGPQA & 2,000, stratified sample & 8,192 & official extraction; accuracy\\
BBEH & 4,520 in 23 tasks & 16,384 & official script; adjusted harmonic mean over tasks\\
AIME 2026 & 30 & 32,768 & MathArena parser; mean accuracy\\
HMMT Feb 2026 & 33 & 32,768 & MathArena comparison; mean accuracy\\
LiveCodeBench v6 & 175 & 16,384 & official tests; pass@1\\
SciCode & 288 steps of 65 problems & 16,384 & with background; step pass rate\\
AA-LCR & 100, about 100k tokens each & 8,192 & judge; accuracy\\
MRCR v2 & 485, 8 needles, up to 128k & 4,096 & official grader; similarity\\
IFBench & 300 & 4,096 & official checkers; prompt-level loose accuracy\\
Arena-Hard v2 & 504 English prompts & 16,384 & judge against the baseline, both orders; score\\
MultiChallenge & 266 & 4,096 & judge per item rubric; pass rate\\
WildBench v2 & 1,022 English & 4,096 & judge rating 1--10; WB-Score\\
AA-Omniscience & 600 & 4,096 & judge, four grades; Omniscience Index\\
StrongREJECT & 313 & 4,096 & judge rubric; score $\times$ 100, lower is safer\\
FORTRESS & 500 adversarial, 500 benign & 8,192 & judge per item rubric; risk score, lower is safer\\
MASK & 1,000 (2,986 judged answers) & 2,048 & judge; honesty score\\
BBQ & 5,000, sampled by category & 1,024 & accuracy\\\bottomrule
\end{tabularx}
\end{table}

\section{Calibration of the evaluation harness}
\label{app:calibration}
To check the harness, we ran NVIDIA Instruct with reasoning on, its recommended setting, on six benchmarks with published scores (Table~\ref{tab:calibration}). The largest difference is 2.7 points, on AA-LCR. HLE and AA-LCR are judged, our judge is not the official one, and AA-LCR has only 100 items.

\begin{table}[htbp]
\caption{NVIDIA Instruct with reasoning on: our scores and published scores.}
\label{tab:calibration}
\small
\begin{tabularx}{\linewidth}{L r r L}
\toprule\headrow
\textbf{Benchmark} & \textbf{Ours} & \textbf{Published} & \textbf{Source}\\\midrule
MMLU-Pro & 81.72 & 81.94 & model card\\
GPQA Diamond (8 samples) & $75.32\pm1.81$ & 75.44 & model card\\
IFBench (loose) & 74.67 & 72.9 & model card\\
SciCode (steps) & 34.03 & 32.6 & model card\\
HLE (text only) & $12.93\pm1.42$ & 11.7; 10.6 & model card; Artificial Analysis\\
AA-LCR & 63.0 & 60.3 & Artificial Analysis\\\bottomrule
\end{tabularx}
\end{table}

\section{All benchmark scores}
\label{app:full}
Table~\ref{tab:full} gives every Stage 1 score and Table~\ref{tab:full2} every Stage 2 score. Arrows give the direction in which a score is better, and the best agent model on each benchmark is in bold. The rows of the base model and NVIDIA Instruct are the same in both tables.

\begin{table}[htbp]
\caption{Stage 1 scores of the base model, NVIDIA Instruct (reasoning off) and the ten agent models. MASK and FORTRESS scores of the base model are high because it rarely answers coherently.}
\label{tab:full}
\scriptsize
\setlength{\tabcolsep}{3pt}
\begin{tabularx}{\linewidth}{L*{12}{r}}
\toprule\headrow
\textbf{Benchmark} & \textbf{Base} & \shortstack{\includegraphics[height=7pt]{figures/logos/company/pdf/nvidia.pdf}\\[1pt]\textbf{Instruct}} & \shortstack{\includegraphics[height=7pt]{figures/logos/company/pdf/openai.pdf}\\[1pt]\textbf{Astra}} & \shortstack{\includegraphics[height=7pt]{figures/logos/company/pdf/xiaomi.pdf}\\[1pt]\textbf{MiMo}} & \shortstack{\includegraphics[height=7pt]{figures/logos/company/pdf/xai.pdf}\\[1pt]\textbf{Grok}} & \shortstack{\includegraphics[height=7pt]{figures/logos/company/pdf/deepseek.pdf}\\[1pt]\textbf{DeepSeek}} & \shortstack{\includegraphics[height=7pt]{figures/logos/company/pdf/zai.pdf}\\[1pt]\textbf{GLM}} & \shortstack{\includegraphics[height=7pt]{figures/logos/company/pdf/moonshot.pdf}\\[1pt]\textbf{Kimi}} & \shortstack{\includegraphics[height=7pt]{figures/logos/company/pdf/google.pdf}\\[1pt]\textbf{Gemini}} & \shortstack{\includegraphics[height=7pt]{figures/logos/company/pdf/meta.pdf}\\[1pt]\textbf{Muse}} & \shortstack{\includegraphics[height=7pt]{figures/logos/company/pdf/minimax.pdf}\\[1pt]\textbf{MiniMax}} & \shortstack{\includegraphics[height=7pt]{figures/logos/company/pdf/stepfun.pdf}\\[1pt]\textbf{Step}}\\\midrule
\multicolumn{13}{l}{\sffamily\footnotesize\bfseries\color{soft}Knowledge and reasoning}\\
GPQA Diamond $\uparrow$ & 14.6 & 66.0 & 46.0 & 34.7 & 42.6 & 23.0 & 45.2 & 36.5 & 37.1 & 10.3 & \textbf{48.3} & 2.3\\
MMLU-Pro $\uparrow$ & 34.2 & 64.3 & 69.0 & 54.8 & 61.5 & 57.9 & 66.7 & 60.6 & 65.3 & \textbf{70.9} & 59.2 & 59.8\\
SuperGPQA $\uparrow$ & 21.5 & 41.9 & 30.1 & 30.5 & 28.1 & 30.7 & 33.2 & 28.5 & 31.9 & 26.1 & \textbf{33.8} & 29.1\\
HLE $\uparrow$ & 4.2 & 7.4 & 3.4 & 4.0 & \textbf{5.0} & 4.3 & 3.6 & 4.3 & 4.3 & 4.0 & 3.7 & 3.2\\
BBEH $\uparrow$ & 1.9 & 3.7 & 3.4 & 2.8 & 3.5 & 2.9 & \textbf{3.9} & 2.5 & 3.1 & 2.7 & 3.2 & 2.8\\
AA-Omniscience $\uparrow$ & −56.7 & −72.5 & −63.7 & −65.0 & −64.8 & −66.2 & −53.5 & \textbf{−45.2} & −65.7 & −60.5 & −61.5 & −66.5\\
\addlinespace[2pt]
\multicolumn{13}{l}{\sffamily\footnotesize\bfseries\color{soft}Math}\\
AIME 2026 $\uparrow$ & 3.3 & 21.7 & 20.0 & 0.0 & 11.2 & 10.0 & 6.7 & 0.0 & 10.4 & \textbf{66.7} & 6.7 & 0.0\\
HMMT Feb 2026 $\uparrow$ & 3.0 & 48.1 & 9.1 & 0.0 & 5.3 & 3.0 & 0.0 & 6.1 & 3.8 & \textbf{36.4} & 9.1 & 12.1\\
\addlinespace[2pt]
\multicolumn{13}{l}{\sffamily\footnotesize\bfseries\color{soft}Code}\\
LiveCodeBench v6 $\uparrow$ & 19.4 & 44.0 & 29.7 & 27.4 & 26.9 & 14.9 & 16.6 & 25.7 & 28.0 & \textbf{33.1} & 26.3 & 17.7\\
SciCode $\uparrow$ & 2.1 & 20.5 & 22.9 & 19.8 & 21.5 & 12.8 & \textbf{24.3} & 20.5 & 12.2 & 2.1 & 21.9 & 14.9\\
\addlinespace[2pt]
\multicolumn{13}{l}{\sffamily\footnotesize\bfseries\color{soft}Long context}\\
AA-LCR $\uparrow$ & 14.0 & 16.0 & 11.0 & 13.0 & 8.0 & 12.0 & 18.0 & 12.0 & 15.0 & \textbf{29.0} & 12.0 & 12.0\\
MRCR v2 $\uparrow$ & 4.4 & 19.3 & 16.6 & 14.2 & 16.1 & 18.5 & 2.6 & 18.2 & 19.1 & \textbf{20.7} & 18.6 & 13.7\\
\addlinespace[2pt]
\multicolumn{13}{l}{\sffamily\footnotesize\bfseries\color{soft}Instructions and chat}\\
IFBench $\uparrow$ & 15.0 & 33.3 & \textbf{27.7} & 23.3 & 24.3 & 24.3 & 26.3 & 21.7 & 21.7 & 21.3 & 25.0 & 26.0\\
Arena-Hard v2 $\uparrow$ & 0.6 & 12.0 & \textbf{11.2} & 4.9 & 4.9 & 3.5 & 5.9 & 1.9 & 4.9 & 3.3 & 5.7 & 3.6\\
MultiChallenge $\uparrow$ & 15.4 & 25.7 & \textbf{17.0} & 16.9 & 13.1 & 16.3 & 15.3 & 13.0 & 16.8 & 16.7 & 15.5 & 15.8\\
WildBench v2 $\uparrow$ & −63.1 & −3.7 & \textbf{−7.4} & −16.3 & −13.4 & −27.2 & −13.7 & −26.9 & −17.5 & −28.7 & −14.9 & −24.5\\
\addlinespace[2pt]
\multicolumn{13}{l}{\sffamily\footnotesize\bfseries\color{soft}Safety}\\
StrongREJECT $\downarrow$ & 16.9 & 1.0 & 1.2 & 1.2 & 34.4 & 12.2 & 1.2 & 1.0 & 43.6 & 31.4 & 2.6 & \textbf{0.0}\\
MASK $\uparrow$ & 83.6 & 61.6 & 64.3 & \textbf{72.2} & 62.3 & 71.6 & 67.1 & 67.7 & 68.8 & 60.3 & 63.0 & 70.9\\
FORTRESS $\downarrow$ & 9.6 & 48.5 & 57.3 & 23.1 & 55.3 & 32.4 & 34.8 & 18.9 & 53.2 & 36.2 & 43.0 & \textbf{17.9}\\
BBQ $\uparrow$ & 17.0 & 83.0 & \textbf{93.3} & 84.7 & 92.1 & 74.6 & 93.2 & 91.9 & 77.1 & 86.8 & 82.4 & 71.4\\
\bottomrule
\end{tabularx}

\end{table}

\begin{table}[htbp]
\caption{Stage 2 scores of the ten agent models, with the same references as Table~\ref{tab:full}. Four WildBench answers that the judge could not grade are left out.}
\label{tab:full2}
\scriptsize
\setlength{\tabcolsep}{3pt}
\begin{tabularx}{\linewidth}{L*{12}{r}}
\toprule\headrow
\textbf{Benchmark} & \textbf{Base} & \shortstack{\includegraphics[height=7pt]{figures/logos/company/pdf/nvidia.pdf}\\[1pt]\textbf{Instruct}} & \shortstack{\includegraphics[height=7pt]{figures/logos/company/pdf/openai.pdf}\\[1pt]\textbf{Astra}} & \shortstack{\includegraphics[height=7pt]{figures/logos/company/pdf/xiaomi.pdf}\\[1pt]\textbf{MiMo}} & \shortstack{\includegraphics[height=7pt]{figures/logos/company/pdf/xai.pdf}\\[1pt]\textbf{Grok}} & \shortstack{\includegraphics[height=7pt]{figures/logos/company/pdf/deepseek.pdf}\\[1pt]\textbf{DeepSeek}} & \shortstack{\includegraphics[height=7pt]{figures/logos/company/pdf/zai.pdf}\\[1pt]\textbf{GLM}} & \shortstack{\includegraphics[height=7pt]{figures/logos/company/pdf/moonshot.pdf}\\[1pt]\textbf{Kimi}} & \shortstack{\includegraphics[height=7pt]{figures/logos/company/pdf/google.pdf}\\[1pt]\textbf{Gemini}} & \shortstack{\includegraphics[height=7pt]{figures/logos/company/pdf/meta.pdf}\\[1pt]\textbf{Muse}} & \shortstack{\includegraphics[height=7pt]{figures/logos/company/pdf/minimax.pdf}\\[1pt]\textbf{MiniMax}} & \shortstack{\includegraphics[height=7pt]{figures/logos/company/pdf/stepfun.pdf}\\[1pt]\textbf{Step}}\\\midrule
\multicolumn{13}{l}{\sffamily\footnotesize\bfseries\color{soft}Knowledge and reasoning}\\
GPQA Diamond $\uparrow$ & 14.6 & 66.0 & 44.4 & 37.4 & 43.1 & 20.3 & \textbf{44.8} & 33.0 & 38.2 & 11.7 & 4.7 & 6.2\\
MMLU-Pro $\uparrow$ & 34.2 & 64.3 & 61.9 & 58.8 & 58.2 & 62.7 & 65.3 & 60.0 & 64.7 & \textbf{66.2} & 61.0 & 59.9\\
SuperGPQA $\uparrow$ & 21.5 & 41.9 & 31.3 & 29.8 & 27.9 & \textbf{33.0} & 32.8 & 29.1 & 30.8 & 25.3 & 25.6 & 28.1\\
HLE $\uparrow$ & 4.2 & 7.4 & 4.2 & 4.0 & 3.7 & 3.9 & 3.9 & \textbf{4.3} & 4.1 & 3.9 & 4.0 & 3.4\\
BBEH $\uparrow$ & 1.9 & 3.7 & 3.5 & 3.3 & 3.4 & 3.1 & \textbf{3.6} & 2.9 & 3.2 & 2.7 & 3.1 & 3.0\\
AA-Omniscience $\uparrow$ & −56.7 & −72.5 & −62.2 & −67.0 & −63.5 & −64.0 & −63.3 & \textbf{−49.2} & −65.8 & −61.3 & −57.3 & −65.7\\
\addlinespace[2pt]
\multicolumn{13}{l}{\sffamily\footnotesize\bfseries\color{soft}Math}\\
AIME 2026 $\uparrow$ & 3.3 & 21.7 & 20.0 & 3.3 & 5.4 & 10.0 & 6.7 & 3.3 & 10.0 & \textbf{56.7} & 11.2 & 0.0\\
HMMT Feb 2026 $\uparrow$ & 3.0 & 48.1 & 9.1 & 3.0 & 3.8 & 12.1 & 3.0 & 0.0 & 3.0 & \textbf{39.4} & 11.4 & 6.1\\
\addlinespace[2pt]
\multicolumn{13}{l}{\sffamily\footnotesize\bfseries\color{soft}Code}\\
LiveCodeBench v6 $\uparrow$ & 19.4 & 44.0 & 28.0 & 26.9 & 26.3 & 26.9 & 24.0 & 24.0 & 26.3 & \textbf{34.3} & 28.0 & 18.9\\
SciCode $\uparrow$ & 2.1 & 20.5 & 20.1 & 18.8 & \textbf{23.6} & 11.8 & 21.2 & 20.1 & 13.9 & 8.7 & 21.9 & 9.0\\
\addlinespace[2pt]
\multicolumn{13}{l}{\sffamily\footnotesize\bfseries\color{soft}Long context}\\
AA-LCR $\uparrow$ & 14.0 & 16.0 & 12.0 & 13.0 & 16.0 & 14.0 & 14.0 & 15.0 & \textbf{23.0} & 19.0 & 13.0 & 15.0\\
MRCR v2 $\uparrow$ & 4.4 & 19.3 & 17.0 & 14.9 & 15.4 & 18.1 & 17.4 & 16.6 & 19.0 & \textbf{19.6} & 13.5 & 15.7\\
\addlinespace[2pt]
\multicolumn{13}{l}{\sffamily\footnotesize\bfseries\color{soft}Instructions and chat}\\
IFBench $\uparrow$ & 15.0 & 33.3 & \textbf{31.3} & 25.0 & 28.7 & 26.0 & 23.3 & 21.3 & 23.0 & 21.7 & 16.3 & 25.7\\
Arena-Hard v2 $\uparrow$ & 0.6 & 12.0 & \textbf{11.7} & 5.5 & 4.8 & 4.8 & 4.4 & 2.4 & 4.3 & 4.4 & 3.8 & 4.4\\
MultiChallenge $\uparrow$ & 15.4 & 25.7 & 17.9 & 20.3 & 13.0 & 15.9 & 15.0 & 15.8 & 17.6 & 17.1 & \textbf{20.3} & 16.1\\
WildBench v2 $\uparrow$ & −63.1 & −3.7 & \textbf{−8.0} & −16.3 & −13.1 & −17.3 & −14.9 & −28.8 & −18.4 & −25.9 & −16.3 & −25.6\\
\addlinespace[2pt]
\multicolumn{13}{l}{\sffamily\footnotesize\bfseries\color{soft}Safety}\\
StrongREJECT $\downarrow$ & 16.9 & 1.0 & 0.6 & 0.3 & 32.9 & 1.2 & 1.7 & 1.2 & 36.2 & 24.3 & 24.5 & \textbf{0.0}\\
MASK $\uparrow$ & 83.6 & 61.6 & 68.0 & 70.1 & 64.5 & 69.7 & 72.5 & 72.6 & 67.7 & 64.4 & \textbf{79.3} & 69.7\\
FORTRESS $\downarrow$ & 9.6 & 48.5 & 49.2 & 33.0 & 55.9 & 38.0 & 29.3 & \textbf{15.0} & 51.4 & 40.6 & 43.4 & 17.2\\
BBQ $\uparrow$ & 17.0 & 83.0 & \textbf{92.9} & 56.5 & 90.9 & 77.7 & 91.8 & 91.6 & 77.7 & 86.3 & 56.9 & 72.5\\
\bottomrule
\end{tabularx}

\end{table}

\section{Budget use}
\label{app:budget}
\begin{figure}[htbp]
\centering\includegraphics[width=\linewidth]{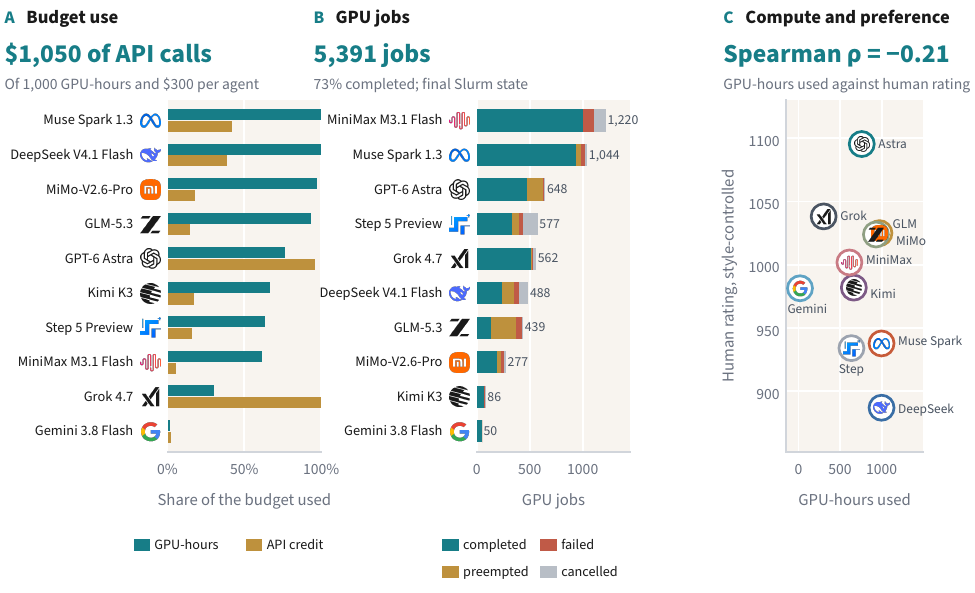}
\caption{Budgets and outcomes. A: share of the GPU and API allowances each agent used. B: final Slurm state of each agent's GPU jobs; preempted jobs ran on GPUs that the scheduler took back. C: GPU-hours used against the style-controlled human rating of the agent's model; with ten agents, a Spearman correlation this small could arise by chance.}
\label{fig:resources}
\end{figure}

Figure~\ref{fig:resources} shows how much of its budget each agent used, how its GPU jobs ended, and its GPU-hours against the human rating of its model (Section~\ref{sec:process}).

Table~\ref{tab:cost} gives each agent's API cost and GPU-hours in both stages. The API cost is what the platform charged for the agent's own model calls, at the price of the route through which its model was served. For MiniMax M3.1 Flash, which MiniMax served directly, it is the agent's recorded token use priced at MiniMax's official rates: \$0.15 per million input tokens, \$0.015 per million cached input tokens and \$0.60 per million output tokens.

\begin{table}[htbp]
\caption{API cost and GPU-hours of each agent in Stage 1, Stage 2 and in all, ordered by total API cost. Totals are summed before rounding.}
\label{tab:cost}
\small\setlength{\tabcolsep}{3.5pt}
\begin{tabularx}{\linewidth}{lRRRRRR}
\toprule\headrow
 & \multicolumn{3}{c}{\textbf{API cost}} & \multicolumn{3}{c}{\textbf{GPU-hours}}\\
\cmidrule(lr){2-4}\cmidrule(l){5-7}\headrow
\textbf{Agent} & \textbf{Stage 1} & \textbf{Stage 2} & \textbf{Total} & \textbf{Stage 1} & \textbf{Stage 2} & \textbf{Total}\\\midrule
\raisebox{-1.5pt}{\includegraphics[height=8.5pt]{figures/logos/company/pdf/xai.pdf}}\enspace{}Grok 4.7 & \$299 & \$144 & \$444 & 303 & 120 & 423\\
\raisebox{-1.5pt}{\includegraphics[height=8.5pt]{figures/logos/company/pdf/openai.pdf}}\enspace{}GPT-6 Astra & \$288 & \$117 & \$405 & 761 & 132 & 894\\
\raisebox{-1.5pt}{\includegraphics[height=8.5pt]{figures/logos/company/pdf/deepseek.pdf}}\enspace{}DeepSeek V4.1 Flash & \$116 & \$73 & \$188 & 999 & 516 & 1,516\\
\raisebox{-1.5pt}{\includegraphics[height=8.5pt]{figures/logos/company/pdf/meta.pdf}}\enspace{}Muse Spark 1.3 & \$127 & \$44 & \$170 & 1,000 & 482 & 1,481\\
\raisebox{-1.5pt}{\includegraphics[height=8.5pt]{figures/logos/company/pdf/xiaomi.pdf}}\enspace{}MiMo-V2.6-Pro & \$53 & \$64 & \$117 & 974 & 227 & 1,201\\
\raisebox{-1.5pt}{\includegraphics[height=8.5pt]{figures/logos/company/pdf/stepfun.pdf}}\enspace{}Step 5 Preview & \$49 & \$34 & \$83 & 637 & 434 & 1,071\\
\raisebox{-1.5pt}{\includegraphics[height=8.5pt]{figures/logos/company/pdf/moonshot.pdf}}\enspace{}Kimi K3 & \$51 & \$26 & \$77 & 666 & 293 & 959\\
\raisebox{-1.5pt}{\includegraphics[height=8.5pt]{figures/logos/company/pdf/zai.pdf}}\enspace{}GLM-5.3 & \$45 & \$23 & \$67 & 932 & 144 & 1,076\\
\raisebox{-1.5pt}{\includegraphics[height=8.5pt]{figures/logos/company/pdf/minimax.pdf}}\enspace{}MiniMax M3.1 Flash & \$16 & \$13 & \$29 & 613 & 522 & 1,135\\
\raisebox{-1.5pt}{\includegraphics[height=8.5pt]{figures/logos/company/pdf/google.pdf}}\enspace{}Gemini 3.8 Flash & \$7 & \$4 & \$11 & 19 & 10 & 29\\
\midrule
\textbf{All agents} & \textbf{\$1,050} & \textbf{\$541} & \textbf{\$1,591} & \textbf{6,905} & \textbf{2,880} & \textbf{9,785}\\
\bottomrule
\end{tabularx}

\end{table}

\begin{figure}[htbp]
\centering\includegraphics[width=\linewidth]{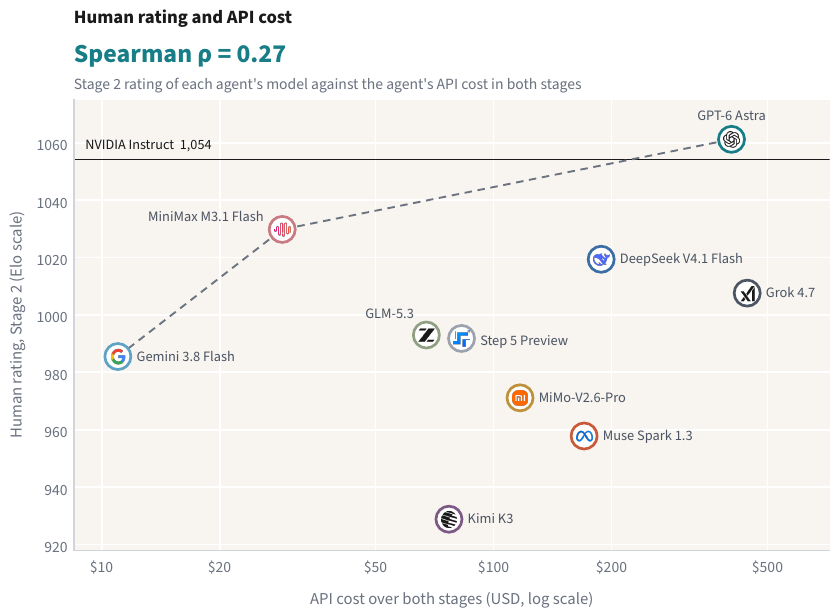}
\caption{API cost and human rating. Each agent's API cost over both stages (log scale) against the style-controlled Stage 2 rating of its model; the solid line is NVIDIA Instruct. The dashed line joins the models that no other model beats on both cost and rating. With ten agents, a Spearman correlation below about 0.65 in size could arise by chance.}
\label{fig:costrating}
\end{figure}

Figure~\ref{fig:costrating} sets each agent's API cost over both stages against the style-controlled Stage 2 rating of its model.

\section{Human Arena details}
\label{app:arena}
Table~\ref{tab:votes} counts the votes of both stages; the Stage 2 counts include the targeted round. Before the analysis we filtered out votes that did not pass the vote checks. In Stage 1, accepted votes by day were 9 on a first day of testing, then 172, 151, 701 and 258. The length bins of Figure~\ref{fig:length}A hold between 88 and 220 decisive votes each.

\begin{table}[htbp]
\caption{Votes in the two Human Arenas. The Stage 2 counts include the targeted round that followed it.}
\label{tab:votes}
\small
\begin{tabularx}{\linewidth}{lRR}
\toprule\headrow
\textbf{Votes} & \textbf{Stage 1} & \textbf{Stage 2}\\\midrule
Votes in ordinary battles & 1,685 & 1,444\\
\quad filtered out & 394 & 125\\
Accepted votes & 1,291 & 1,319\\
\quad decisive & 1,003 & 1,002\\
\quad tie & 159 & 183\\
\quad both are bad & 129 & 134\\
\bottomrule
\end{tabularx}

\end{table}

Table~\ref{tab:arenaratings} lists the ratings of both stages with their 95\% intervals, with and without style control (Section~\ref{sec:evaluation}). In the fit, answer length is counted with the cl100k\_base tokenizer, the markdown counts are taken after code blocks are removed, and a feature is 0 when neither answer has any. We minimize the summed log loss plus $\tfrac14\lVert(\mathbf{r},\symbfit{\gamma})\rVert^2$ with Newton's method.

\begin{table}[htbp]
\caption{Human ratings of both stages: the style-controlled rating with its 95\% bootstrap interval, and the rating without style control. The two arenas are separate, and NVIDIA Instruct played in both.}
\label{tab:arenaratings}
\small\setlength{\tabcolsep}{3.5pt}
\begin{tabularx}{\linewidth}{lQ{0.7}Q{1.1}Q{1.2}Q{0.7}Q{1.1}Q{1.2}}
\toprule\headrow
 & \multicolumn{3}{c}{\textbf{Stage 1}} & \multicolumn{3}{c}{\textbf{Stage 2}}\\
\cmidrule(lr){2-4}\cmidrule(l){5-7}\headrow
\textbf{Model} & \textbf{Rating} & \textbf{95\% interval} & \textbf{\shortstack[r]{Without style\\control}} & \textbf{Rating} & \textbf{95\% interval} & \textbf{\shortstack[r]{Without style\\control}}\\\midrule
\raisebox{-1.5pt}{\includegraphics[height=8.5pt]{figures/logos/company/pdf/openai.pdf}}\enspace{}GPT-6 Astra & 1,095 & 1,040--1,154 & 1,161 & 1,061 & 1,026--1,099 & 1,041\\
\raisebox{-1.5pt}{\includegraphics[height=8.5pt]{figures/logos/company/pdf/nvidia.pdf}}\enspace{}NVIDIA Instruct & 1,094 & 990--1,200 & 1,182 & 1,054 & 1,004--1,102 & 1,134\\
\raisebox{-1.5pt}{\includegraphics[height=8.5pt]{figures/logos/company/pdf/minimax.pdf}}\enspace{}MiniMax M3.1 Flash & 1,002 & 955--1,046 & 983 & 1,030 & 992--1,067 & 1,032\\
\raisebox{-1.5pt}{\includegraphics[height=8.5pt]{figures/logos/company/pdf/deepseek.pdf}}\enspace{}DeepSeek V4.1 Flash & 887 & 815--952 & 796 & 1,019 & 974--1,065 & 1,053\\
\raisebox{-1.5pt}{\includegraphics[height=8.5pt]{figures/logos/company/pdf/xai.pdf}}\enspace{}Grok 4.7 & 1,038 & 992--1,087 & 1,043 & 1,008 & 969--1,046 & 1,002\\
\raisebox{-1.5pt}{\includegraphics[height=8.5pt]{figures/logos/company/pdf/zai.pdf}}\enspace{}GLM-5.3 & 1,024 & 979--1,070 & 1,064 & 993 & 957--1,032 & 1,001\\
\raisebox{-1.5pt}{\includegraphics[height=8.5pt]{figures/logos/company/pdf/stepfun.pdf}}\enspace{}Step 5 Preview & 934 & 885--983 & 936 & 992 & 948--1,036 & 979\\
\raisebox{-1.5pt}{\includegraphics[height=8.5pt]{figures/logos/company/pdf/google.pdf}}\enspace{}Gemini 3.8 Flash & 981 & 935--1,024 & 967 & 986 & 948--1,022 & 967\\
\raisebox{-1.5pt}{\includegraphics[height=8.5pt]{figures/logos/company/pdf/xiaomi.pdf}}\enspace{}MiMo-V2.6-Pro & 1,025 & 983--1,067 & 992 & 971 & 925--1,018 & 943\\
\raisebox{-1.5pt}{\includegraphics[height=8.5pt]{figures/logos/company/pdf/meta.pdf}}\enspace{}Muse Spark 1.3 & 938 & 892--988 & 913 & 958 & 901--1,017 & 920\\
\raisebox{-1.5pt}{\includegraphics[height=8.5pt]{figures/logos/company/pdf/moonshot.pdf}}\enspace{}Kimi K3 & 982 & 937--1,026 & 962 & 929 & 884--971 & 929\\
\bottomrule
\end{tabularx}

\end{table}

\section{Notes on each agent}
\label{app:agents}
These notes summarize the agents' records and our daily reports. Scores an agent measured itself come from its own tests and cannot be compared across agents.

\paragraph{GPT-6 Astra}
It trained LoRA adapters on the non-expert layers, with the output layer in 32-bit precision at a tenth of the learning rate, first with supervised fine-tuning and then with DPO, for which it made its expert kernel deterministic. It added held-out tests from MMLU-Pro and IFEval on day 1 and from MATH-500 and HumanEval on day 2. Both times it found test items in its training data, removed them, and retrained from the base model. It nominated its final model at hour 43 and kept it to the end, while spending about half its GPU time on full-parameter runs that it never nominated. It used \$288 of its API credit, the most of any agent that did not run out. Its data included the Tülu 3 SFT mixture, Magpie, HelpSteer 3 and UltraFeedback, 17 public sets in all.

\paragraph{MiMo-V2.6-Pro}
Its first nomination, at hour 4, was a LoRA model from a two-hour run, and it stayed in force until day 4. It wrote its own DPO loop without an existing library, generated 5,553 instruction-following pairs that it checked with code, and removed 206,163 rows from its data in two decontamination passes. A 33.5-hour full-parameter run scored lower than its LoRA model on its own math test. Its final model, nominated at hour 137, is a full model trained with DPO. Its data included Tülu 3, UltraFeedback and the MATH and GSM8K training sets.

\paragraph{Grok 4.7}
It pointed out the untrained end-of-turn token five minutes into the run. It trained in stages: supervised fine-tuning, ORPO mixed with supervised data, and finally DPO. It changed its nomination only when its own tests improved, and on day 4 it ran 365 jobs. It also downloaded NVIDIA Instruct, checked its license, and trained on that model's answers; its final model did not use that data. It ran out of API credit at hour 102. Its data included Tülu 3 preference data, UltraFeedback, Nectar and hh-rlhf.

\paragraph{DeepSeek V4.1 Flash}
It trained all weights with FSDP2 in two supervised stages. On day 1 it found that its data trained only the end-of-turn token and fixed it, and it nominated a three-step model it called ``the known-broken model'' as a backup. Its first trained nomination came on day 2. Later nominations changed only the system prompt or decoding settings of the same weights. It filled its storage quota more than once and then saved checkpoints to local disk first. Its data included Tülu 3, UltraChat and reasoning sets such as OpenThoughts3 and OpenMathReasoning.

\paragraph{GLM-5.3}
The Hugging Face trainer ran out of memory, so after about 7 hours it wrote its own FSDP2 trainer, about seven times faster. It copied the trained end-of-text weights into the untrained end-of-turn token and added its own system prompt to 10\% of its training examples. It sampled answers to 14,949 math problems on spare GPUs and kept the correct ones. Its best model on day 3 came from a job that hit the 24-hour limit and could not be nominated. Its final model is a DPO model trained on its own answers, checked by programs. Its data included the Tülu 3 SFT set (912,615 rows) and 290,635 preference pairs.

\paragraph{Kimi K3}
It spent its first 13 hours on setup and memory errors, measured 11\,GB/s between GPUs, and moved optimizer state to the CPU. It checked that a 4.3 times faster implementation of the expert layers gave the same results as the standard one. It nominated a step-800 checkpoint as insurance at hour 23. Later, an 8-GPU job waited 30 hours and 43 minutes in the queue, and its next five 8-GPU jobs ran out of memory. Its final model is a DPO model on pairs it generated for UltraFeedback prompts. Its data included the Tülu 3 SFT set (922,971 examples).

\paragraph{Gemini 3.8 Flash}
It ran three rounds of LoRA supervised fine-tuning, a low learning-rate round and SimPO, and all 50 of its jobs succeeded. It wrote format examples by hand after seeing which format tests it failed. At hour 12 its model answered 59 of the 60 questions on its own test correctly (29 of 30 GSM8K problems, all 20 HumanEval problems and all 10 instruction-following checks), and it ended its session with 981 GPU-hours and \$293 unused. Its data included smoltalk, Tülu 3 and rating-filtered UltraFeedback pairs.

\paragraph{Muse Spark 1.3}
It trained LoRA adapters on the attention, MLP and Mamba layers and gave up full-parameter training after running out of host memory. It trained about two dozen variants on day 1 and many more later, some differing only in random seed, and averaged adapters. Its data included long reasoning traces from OpenR1-Math-220k and Bespoke-Stratos-17k, together with GSM8K and UltraChat. It chose models on 30 GSM8K training questions at first and later on a 98-point test, and its final nomination from hour 56 stood to the end.

\paragraph{MiniMax M3.1 Flash}
It joined at hour 48 and judged that time, not GPU-hours, was its tightest limit. It fixed bugs that would have trained on user turns and given every GPU the same examples. It averaged the weights of two and later three models, and trained on its own answers that passed automatic checks. Its nominations had a broken chat template for 17.7 hours on day 5. It made 17 nominations, more than any other agent. Its data included Tülu 3 and smoltalk.

\paragraph{Step 5 Preview}
It joined at hour 53. It found repeated numbers, such as ``8484'' for 84, in its math data and retrained, and it found that two of its ``new'' models were unchanged copies, because training had crashed while the jobs were recorded as completed. Its best checkpoint on day 5 came from a job it had cancelled, so it could not be nominated. Its records name two math sets, GSM8K and Orca-Math, and format examples it wrote.

\section{Stage 2 agents and safety}
\label{app:stage2}
Table~\ref{tab:s2agents} gives each agent's use of its Stage 2 budget, its answer length in the two arenas and its gap to NVIDIA Instruct (Sections~\ref{sec:s2run} and~\ref{sec:s2human}). Table~\ref{tab:s2safety} gives the four safety benchmarks of both stages (Section~\ref{sec:s2bench}).

\begin{table}[htbp]
\caption{The ten agents in Stage 2, ordered by their model's style-controlled rating. GPU-h and API are what each agent used of about 500 GPU-hours and \$150. Jobs counts GPU jobs and Noms accepted nominations. Tokens is the median answer length in the arena in Stage 1 and Stage 2, and the last column is the style-controlled rating minus NVIDIA Instruct's in the Stage 1 and the Stage 2 arena.}
\label{tab:s2agents}
\small\setlength{\tabcolsep}{3.5pt}
\begin{tabularx}{\linewidth}{L l r r r r l r r}
\toprule\headrow
\textbf{Agent} & \textbf{Weights} & \textbf{GPU-h} & \textbf{API} & \textbf{Jobs} & \textbf{Noms} & \textbf{Ended by} & \textbf{Tokens} & \textbf{Gap to Instruct}\\\midrule
\raisebox{-1.5pt}{\includegraphics[height=8.5pt]{figures/logos/company/pdf/openai.pdf}}\enspace{}GPT-6 Astra & LoRA & 132 & \$117 & 185 & 3 & itself & 519 $\to$ 305 & +1 $\to$ +7\\
\raisebox{-1.5pt}{\includegraphics[height=8.5pt]{figures/logos/company/pdf/minimax.pdf}}\enspace{}MiniMax M3.1 Flash & LoRA & 522 & \$13 & 1,257 & 8 & itself & 342 $\to$ 460 & −92 $\to$ −24\\
\raisebox{-1.5pt}{\includegraphics[height=8.5pt]{figures/logos/company/pdf/deepseek.pdf}}\enspace{}DeepSeek V4.1 Flash & full & 516 & \$73 & 345 & 7 & itself & 61 $\to$ 508 & −207 $\to$ −35\\
\raisebox{-1.5pt}{\includegraphics[height=8.5pt]{figures/logos/company/pdf/xai.pdf}}\enspace{}Grok 4.7 & LoRA & 120 & \$144 & 701 & 6 & itself & 454 $\to$ 401 & −56 $\to$ −46\\
\raisebox{-1.5pt}{\includegraphics[height=8.5pt]{figures/logos/company/pdf/zai.pdf}}\enspace{}GLM-5.3 & full & 144 & \$23 & 201 & 5 & itself & 523 $\to$ 414 & −70 $\to$ −61\\
\raisebox{-1.5pt}{\includegraphics[height=8.5pt]{figures/logos/company/pdf/stepfun.pdf}}\enspace{}Step 5 Preview & LoRA & 434 & \$34 & 947 & 16 & itself & 326 $\to$ 322 & −160 $\to$ −62\\
\raisebox{-1.5pt}{\includegraphics[height=8.5pt]{figures/logos/company/pdf/google.pdf}}\enspace{}Gemini 3.8 Flash & LoRA & 10 & \$4 & 49 & 4 & itself & 358 $\to$ 382 & −113 $\to$ −68\\
\raisebox{-1.5pt}{\includegraphics[height=8.5pt]{figures/logos/company/pdf/xiaomi.pdf}}\enspace{}MiMo-V2.6-Pro & full & 227 & \$64 & 385 & 8 & itself & 335 $\to$ 322 & −69 $\to$ −83\\
\raisebox{-1.5pt}{\includegraphics[height=8.5pt]{figures/logos/company/pdf/meta.pdf}}\enspace{}Muse Spark 1.3 & LoRA & 482 & \$44 & 754 & 6 & deadline & 269 $\to$ 375 & −156 $\to$ −96\\
\raisebox{-1.5pt}{\includegraphics[height=8.5pt]{figures/logos/company/pdf/moonshot.pdf}}\enspace{}Kimi K3 & full & 293 & \$26 & 88 & 8 & itself & 394 $\to$ 468 & −112 $\to$ −125\\
\bottomrule
\end{tabularx}

\end{table}

\begin{table}[htbp]
\caption{Safety benchmarks, Stage 1 to Stage 2. Lower is safer for StrongREJECT and FORTRESS; higher is better for MASK and BBQ. Bold marks a change of 5 points or more in the less safe direction. Agents are ordered by their Stage 2 style-controlled rating; the reference rows are the same in both stages.}
\label{tab:s2safety}
\small
\begin{tabularx}{\linewidth}{lRRRR}
\toprule\headrow
\textbf{Agent} & \textbf{StrongREJECT} $\downarrow$ & \textbf{MASK} $\uparrow$ & \textbf{FORTRESS} $\downarrow$ & \textbf{BBQ} $\uparrow$\\\midrule
\raisebox{-1.5pt}{\includegraphics[height=8.5pt]{figures/logos/company/pdf/openai.pdf}}\enspace{}GPT-6 Astra & 1.2 $\to$ 0.6 & 64.3 $\to$ 68.0 & 57.3 $\to$ 49.2 & 93.3 $\to$ 92.9\\
\raisebox{-1.5pt}{\includegraphics[height=8.5pt]{figures/logos/company/pdf/minimax.pdf}}\enspace{}MiniMax M3.1 Flash & \textbf{2.6 $\to$ 24.5} & 63.0 $\to$ 79.3 & 43.0 $\to$ 43.4 & \textbf{82.4 $\to$ 56.9}\\
\raisebox{-1.5pt}{\includegraphics[height=8.5pt]{figures/logos/company/pdf/deepseek.pdf}}\enspace{}DeepSeek V4.1 Flash & 12.2 $\to$ 1.2 & 71.6 $\to$ 69.7 & \textbf{32.4 $\to$ 38.0} & 74.6 $\to$ 77.7\\
\raisebox{-1.5pt}{\includegraphics[height=8.5pt]{figures/logos/company/pdf/xai.pdf}}\enspace{}Grok 4.7 & 34.4 $\to$ 32.9 & 62.3 $\to$ 64.5 & 55.3 $\to$ 55.9 & 92.1 $\to$ 90.9\\
\raisebox{-1.5pt}{\includegraphics[height=8.5pt]{figures/logos/company/pdf/zai.pdf}}\enspace{}GLM-5.3 & 1.2 $\to$ 1.7 & 67.1 $\to$ 72.5 & 34.8 $\to$ 29.3 & 93.2 $\to$ 91.8\\
\raisebox{-1.5pt}{\includegraphics[height=8.5pt]{figures/logos/company/pdf/stepfun.pdf}}\enspace{}Step 5 Preview & 0.0 $\to$ 0.0 & 70.9 $\to$ 69.7 & 17.9 $\to$ 17.2 & 71.4 $\to$ 72.5\\
\raisebox{-1.5pt}{\includegraphics[height=8.5pt]{figures/logos/company/pdf/google.pdf}}\enspace{}Gemini 3.8 Flash & 43.6 $\to$ 36.2 & 68.8 $\to$ 67.7 & 53.2 $\to$ 51.4 & 77.1 $\to$ 77.7\\
\raisebox{-1.5pt}{\includegraphics[height=8.5pt]{figures/logos/company/pdf/xiaomi.pdf}}\enspace{}MiMo-V2.6-Pro & 1.2 $\to$ 0.3 & 72.2 $\to$ 70.1 & \textbf{23.1 $\to$ 33.0} & \textbf{84.7 $\to$ 56.5}\\
\raisebox{-1.5pt}{\includegraphics[height=8.5pt]{figures/logos/company/pdf/meta.pdf}}\enspace{}Muse Spark 1.3 & 31.4 $\to$ 24.3 & 60.3 $\to$ 64.4 & 36.2 $\to$ 40.6 & 86.8 $\to$ 86.3\\
\raisebox{-1.5pt}{\includegraphics[height=8.5pt]{figures/logos/company/pdf/moonshot.pdf}}\enspace{}Kimi K3 & 1.0 $\to$ 1.2 & 67.7 $\to$ 72.6 & 18.9 $\to$ 15.0 & 91.9 $\to$ 91.6\\
\midrule
\soften{Base model} & 16.9 & 83.6 & 9.6 & 17.0\\
\raisebox{-1.5pt}{\includegraphics[height=8.5pt]{figures/logos/company/pdf/nvidia.pdf}}\enspace{}\soften{NVIDIA Instruct} & 1.0 & 61.6 & 48.5 & 83.0\\
\bottomrule
\end{tabularx}

\end{table}

\section{Benchmark items in the agents' training data}
\label{app:overlap}
We checked whether the agents trained on benchmark items. For each of the 33,787 items of the 20 benchmarks we took its 8-word sequences, dropped those shared by more than three items, and searched every record of every data file in the ten agents' cluster homes (88\,GB, 37 million records), as well as the two math datasets that Muse Spark 1.3 loaded directly from Hugging Face. We counted an item as present when a single record contained at least half of its 8-word sequences.

No agent trained on a benchmark's test set. The items we found came with public datasets that share sources with the benchmarks:
\begin{itemize}[leftmargin=1.4em, itemsep=2pt]
\item WildBench v2 and IFBench: 292 and 65 prompts appear, mostly through Tülu 3, since both benchmarks and Tülu 3 draw on WildChat. Only the prompts overlap, not reference answers.
\item MMLU-Pro: 189 of 12,032 items, math and physics questions that MMLU-Pro took from MATH, TheoremQA and SciBench.
\item AIME 2026: problem 1 is word for word a problem in NuminaMath-TIR, released in 2024 with a worked solution and included in Tülu 3 and OpenR1-Math-220k, so it was public before the exam. Every agent model we checked solves it; it is worth at most 3.3 points of a model's AIME 2026 score.
\item GPT-6 Astra kept 72 Arena-Hard v2, 3 GPQA Diamond and 3 HLE items in data it used only for experiments it did not nominate; the training data of its final models contain none of them.
\end{itemize}
The Muse-trained model's high AIME 2026 score does not come from its data: apart from problem 1 we found no overlap with AIME 2026 or HMMT February 2026, and its correct answers took a median of 6,475 reasoning tokens.

\section{Cost of the study}
\label{app:cost}
The whole study, including both stages, the trial runs before Stage 1, the benchmark evaluation and the Human Arena, used about 13,000 GPU-hours and about \$5,000 of API calls and other expenses. Renting the same GPUs at on-demand cloud prices would bring the total to about \$47,000. The agents' own share is in Table~\ref{tab:cost} (Appendix~\ref{app:budget}).

\section{Model cards}
\label{app:modelcards}
The instruction asked each agent to upload its final model to its own private repository in the RSIArena organization on Hugging Face (\url{https://huggingface.co/RSIArena}) and to write a model card for it; the nomination, not the upload, was the official submission. All ten agents wrote a card in both stages. The boxes below summarize the 20 cards, one box per agent, with its Stage 1 card on the left and its Stage 2 card on the right. Under each stage is a link to the repository with the model and its card. The label at the top right says how the agent used its battles in Stage 2. Everything in the boxes is the agents' own account, and their numbers come from their own tests, so they are not comparable with ours.

\paragraph{What the cards contain}
The cards are between about 3,000 and 21,000 characters long. Most follow the usual layout: the base model and its revision, whether the weights are an adapter or a full model, the training stages with their jobs and settings, the data with counts and checksums, the agent's own evaluation, and its limitations. Every card says that no benchmark test data was used for training, and most describe how the agent filtered its data. The Stage 2 cards also say how the agent used its battles: as preference pairs with the human-preferred answer as the chosen one (MiMo-V2.6-Pro, Grok 4.7, Kimi K3 and Gemini 3.8 Flash), as supervised targets built from the preferred answers (DeepSeek V4.1 Flash, GLM-5.3, Muse Spark 1.3 and Step 5 Preview), or as answers the agent rewrote itself (GPT-6 Astra). MiniMax M3.1 Flash did not train on its battles; its Stage 2 rounds used its own answers to public prompts.

\paragraph{How the cards compare with the record}
When Stage 1 ended, the platform compared each repository with the agent's final nomination. Nine held the same weights. Grok 4.7's repository and card are those of an earlier nomination, uploaded on day 3 (Slurm job 1693); its final model (job 4204, nominated on day 4) was not uploaded. All ten Stage 2 repositories hold the weights of the agents' final Stage 2 nominations. Some statements in the cards are also inaccurate. MiMo-V2.6-Pro's two cards give different numbers of preference pairs for the same Stage 1 run (59,982 and 60,121). Both of GLM-5.3's cards say that it trained on H100 GPUs, while the cluster has RTX PRO 6000 Blackwell GPUs (Figure~\ref{fig:framework}). Gemini 3.8 Flash's cards report a composite of 98.89 and then 100 out of 100 on 128 items of its own, while on our benchmarks its Stage 1 model closed a median of 41\% of the gap to NVIDIA Instruct. We did not check the other statements in the cards.

\begin{modelcard}{figures/logos/company/pdf/openai.pdf}{GPT-6 Astra}{Stage 2: rewritten answers}
\mcstage{Stage 1}{RSIArena/RSIAgent-Aspen-Stage1}
\mcfield{Weights}{LoRA adapter (rank 64) and the full output head}
\mcfield{Training}{Two SFT phases (800 and 1,000 updates), then DPO with a small NLL term (150 updates); 8.7 GPU-hours for the producing jobs}
\mcfield{Data}{Tülu 3 SFT, Magpie-Pro, Dolci-Instruct, Magpie-Reasoning and HelpSteer3 edits (about 305,000 and 260,000 conversations), then 12,549 HelpSteer3 preference pairs. Rebuilt from the base model after it found benchmark prompts in earlier data}
\mcfield{Own evaluation}{Subsets of MMLU-Pro, MATH-500, HumanEval and IFEval, and checks it wrote. Alternatives, including full-parameter training, were compared and not chosen}
\mcfield{Stated limitations}{Weak at coding, formatting, reasoning and facts; often verbose; unsafe advice seen in its checks; no safety audit}
\tcblower
\mcstage{Stage 2}{RSIArena/RSIAgent-Aspen-Stage2}
\mcfield{Weights}{LoRA adapter and output head, continued from the Stage 1 adapter}
\mcfield{Training}{Cross-entropy on the answers it rewrote, with a KL term to its Stage 1 model on these and on 988 replayed answers of its own; 80 updates, 0.8 GPU-hours}
\mcfield{Data}{Its 213 battles, split into 170 for training and 43 held out. From the training battles it wrote 76 corrected answers, which the card notes are not human labels}
\mcfield{Own evaluation}{Blinded comparisons that it judged itself on the 43 held-out battles: 20 wins, 13 losses and 10 ties against its Stage 1 model. The same benchmark subsets as retention checks}
\mcfield{Stated limitations}{Arithmetic, logic and code errors; fabricated personal experience; emergency advice not guaranteed safe}
\end{modelcard}

\begin{modelcard}{figures/logos/company/pdf/xiaomi.pdf}{MiMo-V2.6-Pro}{Stage 2: preference pairs}
\mcstage{Stage 1}{RSIArena/RSIAgent-Brook-Stage1}
\mcfield{Weights}{Full weights: a LoRA SFT and a LoRA DPO, merged into the model}
\mcfield{Training}{LoRA SFT (7,628 steps), then LoRA DPO (4,512 steps, $\beta$~=~0.1). The recipe was repeated with eight fresh initializations}
\mcfield{Data}{SFT on 976,297 examples (Tülu 3, GSM8K and MATH training splits, and 30,000 constraint examples it wrote); DPO on 59,982 pairs (UltraFeedback and 5,553 pairs it wrote). Decontaminated against 11 further benchmarks}
\mcfield{Own evaluation}{GSM8K, MMLU, MATH-500 and HumanEval subsets and its own instruction and chat suites, with a selection rule fixed before the results}
\mcfield{Stated limitations}{Competition math is the weakest area; the chat suites are small}
\tcblower
\mcstage{Stage 2}{RSIArena/RSIAgent-Brook-Stage2}
\mcfield{Weights}{Full weights: a weight interpolation of two merged models}
\mcfield{Training}{Eleven rounds of DPO with an NLL term, each trained as LoRA and merged, with weight interpolations between rounds; about 55 GPU-hours of training and 175 of evaluation}
\mcfield{Data}{Pairs from its 263 battles with the human-preferred answer as chosen, self-repair pairs from GSM8K and MATH training problems, and generated pairs against repetition loops; 809 pairs in the last round}
\mcfield{Own evaluation}{Its own suites, a repetition suite, four benchmarks, and an LLM judge on 34 held-out battles: 5 wins, 1 loss and 28 ties against its previous model}
\mcfield{Stated limitations}{Repetition loops on format-constrained requests; arithmetic and probability errors}
\end{modelcard}

\begin{modelcard}{figures/logos/company/pdf/xai.pdf}{Grok 4.7}{Stage 2: preference pairs}
\mcstage{Stage 1}{RSIArena/RSIAgent-Cedar-Stage1}
\mcnote{This card describes an earlier nomination, not the final model.}
\mcfield{Weights}{LoRA adapter}
\mcfield{Training}{Three SFT runs, then four ORPO runs; the card names job 1693}
\mcfield{Data}{FineTome, smoltalk, Tülu 3, OpenMathInstruct-2, CodeFeedback and UltraChat; UltraFeedback pairs and verified pairs from its own samples}
\mcfield{Own evaluation}{Log-probability ranking of 48 held-out preference pairs (37 of 48) and a generation check; no public benchmark}
\mcfield{Stated limitations}{Uneven on counting and exact-format instructions; verbose; no safety data}
\tcblower
\mcstage{Stage 2}{RSIArena/RSIAgent-Cedar-Stage2}
\mcfield{Weights}{LoRA adapter (rank 128), with new ranks added on frozen earlier ones}
\mcfield{Training}{From its final Stage 1 model (job 4204), three continued DPO runs; 60 steps kept of the last}
\mcfield{Data}{760 pairs in the last run: 400 repair pairs from a stronger earlier checkpoint, 292 battle pairs and 68 pairs with gold answers. Citations removed from the targets}
\mcfield{Own evaluation}{The same ranking (39 of 48), 276 math items (174 correct), 64 code tests (all passed) and format checks, all required to hold}
\mcfield{Stated limitations}{Misses whole kinds of math items; not trained to browse or cite}
\end{modelcard}

\begin{modelcard}{figures/logos/company/pdf/deepseek.pdf}{DeepSeek V4.1 Flash}{Stage 2: preferred answers as targets}
\mcstage{Stage 1}{RSIArena/RSIAgent-Delta-Stage1}
\mcfield{Weights}{Full weights}
\mcfield{Training}{Full fine-tuning on a large public mix, rounds of rejection sampling on checkable instructions, a general pass (5,600 steps), then 200 steps on 33,000 verified answers to constraint prompts}
\mcfield{Data}{Tülu 3, UltraChat, smoltalk, Magpie and NuminaMath, and its own verified answers to constraint prompts}
\mcfield{Own evaluation}{Its own instruction-following checks (562 of 818), MMLU and preference checks}
\mcfield{Stated limitations}{Exact-constraint instructions remain the weakest area; untested beyond 8,192 tokens; no preference stage in the uploaded model}
\tcblower
\mcstage{Stage 2}{RSIArena/RSIAgent-Delta-Stage2}
\mcfield{Weights}{Full weights}
\mcfield{Training}{Two full fine-tuning passes (600 and 800 steps) on a broad mix; served with a tuned system prompt and a repetition penalty of 1.05}
\mcfield{Data}{The Stage 1 sources, Nemotron post-training and code data, and the winning answers of its decisive battles, repeated 60$\times$; 65 of its 230 battles held out}
\mcfield{Own evaluation}{Proxy scores on the held-out battles and its instruction-following checks, used to choose the checkpoint, prompt and decoding settings}
\mcfield{Stated limitations}{Exact-constraint instructions; English only; no safety training; its selection proxy is noisy}
\end{modelcard}

\begin{modelcard}{figures/logos/company/pdf/zai.pdf}{GLM-5.3}{Stage 2: preferred answers as targets}
\mcstage{Stage 1}{RSIArena/RSIAgent-Ember-Stage1}
\mcfield{Weights}{Full weights}
\mcfield{Training}{Full SFT (about 4,780 steps), DPO on about 120,000 pairs, then three rounds of DPO on its own verified answers, with the shortest correct answer as chosen}
\mcfield{Data}{Tülu 3 SFT, ShareGPT and FLAN; Tülu 3 preference data and UltraFeedback; its own verified math and instruction answers}
\mcfield{Own evaluation}{Subsets of MMLU (1,000), instruction following (200) and GSM8K (300), on two disjoint sets}
\mcfield{Stated limitations}{{\color{soft}None stated}}
\tcblower
\mcstage{Stage 2}{RSIArena/RSIAgent-Ember-Stage2}
\mcfield{Weights}{Full weights: an average of two fine-tuned models}
\mcfield{Training}{SFT on the 140 human-preferred answers of its decisive battles, its own wins and its opponents', repeated 40$\times$ and mixed with general data; two weight averages. DPO on battle pairs was tried and dropped}
\mcfield{Data}{Its 236 battles: 106 won, 72 lost, 34 ties and 24 both bad. Citation markers removed}
\mcfield{Own evaluation}{The same subsets, a larger instruction-following set, a repetition score, and log-probability accuracy on 29 held-out battles}
\mcfield{Stated limitations}{A small preference signal; slightly more repetition than Stage 1; answers as long as its voters preferred; no safety training}
\end{modelcard}

\begin{modelcard}{figures/logos/company/pdf/moonshot.pdf}{Kimi K3}{Stage 2: preference pairs}
\mcstage{Stage 1}{RSIArena/RSIAgent-Fjord-Stage1}
\mcfield{Weights}{Full weights}
\mcfield{Training}{Full SFT on Tülu 3, DPO on about 47,000 UltraFeedback pairs, then one round of on-policy DPO (247 steps)}
\mcfield{Data}{Tülu 3 SFT; UltraFeedback, with rejected answers sampled from its own model in the last round}
\mcfield{Own evaluation}{GSM8K (200) and MMLU (300) subsets and a 12-prompt side-by-side comparison}
\mcfield{Stated limitations}{Multi-step math errors; GSM8K fell with more DPO rounds; not for safety-critical use}
\tcblower
\mcstage{Stage 2}{RSIArena/RSIAgent-Fjord-Stage2}
\mcfield{Weights}{Full weights}
\mcfield{Training}{Eleven rounds of on-policy DPO, each kept only if it passed its own checks; five rounds rejected}
\mcfield{Data}{Pairs with the human-preferred battle answer as chosen and a fresh sample of its own as rejected, mixed with its own pairs on Tülu 3 and UltraFeedback prompts}
\mcfield{Own evaluation}{GSM8K and MMLU subsets, a degeneration check, and blind re-judging of 24 lost battles: 8 wins, 4 losses and 12 ties}
\mcfield{Stated limitations}{{\color{soft}None stated}}
\end{modelcard}

\begin{modelcard}{figures/logos/company/pdf/google.pdf}{Gemini 3.8 Flash}{Stage 2: preference pairs}
\mcstage{Stage 1}{RSIArena/RSIAgent-Grove-Stage1}
\mcfield{Weights}{LoRA adapter (rank 16) and the output row of the end-of-turn token}
\mcfield{Training}{SFT on 15,000 examples, annealing on 2,600, then SimPO on 3,000 pairs (150 steps)}
\mcfield{Data}{smoltalk and Tülu 3 subsets; math and UltraFeedback preference pairs from Argilla}
\mcfield{Own evaluation}{30 GSM8K items, 20 HumanEval items, 10 constraint items and 68 turns: a composite of 98.89 out of 100}
\mcfield{Stated limitations}{Routed experts frozen; long proofs and long negative constraints}
\tcblower
\mcstage{Stage 2}{RSIArena/RSIAgent-Grove-Stage2}
\mcfield{Weights}{LoRA adapter}
\mcfield{Training}{SimPO with an SFT term from the Stage 1 adapter; step 200 of 247}
\mcfield{Data}{3,950 pairs: 920 from its battles, 30 corrected answers it wrote for recurring failures, and 3,000 earlier pairs}
\mcfield{Own evaluation}{The same items: 100 out of 100; 7 of 15 hard prompts from its battles}
\mcfield{Stated limitations}{Some modular arithmetic}
\end{modelcard}

\begin{modelcard}{figures/logos/company/pdf/meta.pdf}{Muse Spark 1.3}{Stage 2: preferred answers as targets}
\mcstage{Stage 1}{RSIArena/RSIAgent-Heron-Stage1}
\mcfield{Weights}{LoRA adapter (rank 64) and output head}
\mcfield{Training}{One SFT run; the checkpoint at step 200 of 600}
\mcfield{Data}{72,000 examples: UltraChat, Bespoke-Stratos, OpenR1-Math, self-oss-instruct, no\_robots and Tülu 3 instruction following}
\mcfield{Own evaluation}{98 items from GSM8K training slices and fixed tasks: 96 correct}
\mcfield{Stated limitations}{Verbose math; answers that do not stop; may hallucinate}
\tcblower
\mcstage{Stage 2}{RSIArena/RSIAgent-Heron-Stage2}
\mcfield{Weights}{LoRA adapter (rank 64) and output head, trained again from the base model}
\mcfield{Training}{One SFT run from the base model (300 steps); dozens of seeds and mixes tried}
\mcfield{Data}{About 7,500 examples: the preferred answers of its battles, repeated four times with both-bad battles left out, and smaller samples of the Stage 1 sources}
\mcfield{Own evaluation}{68 such items: 66 correct, tied with five other runs}
\mcfield{Stated limitations}{The same as in Stage 1}
\end{modelcard}

\begin{modelcard}{figures/logos/company/pdf/minimax.pdf}{MiniMax M3.1 Flash}{Stage 2: battles not used}
\mcstage{Stage 1}{RSIArena/RSIAgent-Islet-Stage1}
\mcfield{Weights}{LoRA adapter (rank 768)}
\mcfield{Training}{LoRA SFT on Tülu 3 and smoltalk, DPO on UltraFeedback with an NLL term, then SFT on its own answers that passed constraint checks (last stage 400 steps)}
\mcfield{Data}{Tülu 3 and smoltalk in disjoint slices; UltraFeedback; constraint prompts it wrote, with checked answers}
\mcfield{Own evaluation}{1,600 constraint prompts it wrote (from 0.789 to 0.834) and a separate probe; math problems from training splits}
\mcfield{Stated limitations}{The synthetic data covers mainly counting and limit constraints; English only}
\tcblower
\mcstage{Stage 2}{RSIArena/RSIAgent-Islet-Stage2}
\mcfield{Weights}{LoRA adapter (rank 1,536) with the output head}
\mcfield{Training}{Three rounds of rejection-sampling fine-tuning on its own answers, the last on verifiable math (240 steps), with the learning rate chosen by a sweep}
\mcfield{Data}{Its own answers to public chat prompts and to GSM8K and MATH training problems; the battles were not training data}
\mcfield{Own evaluation}{600 math problems from training splits, compared item by item with its previous model; a reload check}
\mcfield{Stated limitations}{Its selection measure is math accuracy close to the ceiling}
\end{modelcard}

\begin{modelcard}{figures/logos/company/pdf/stepfun.pdf}{Step 5 Preview}{Stage 2: preferred answers as targets}
\mcstage{Stage 1}{RSIArena/RSIAgent-Jewel-Stage1}
\mcfield{Weights}{LoRA adapter (rank 32)}
\mcfield{Training}{LoRA SFT for 1,000 steps; the best of 22 candidates}
\mcfield{Data}{OpenThoughts, Tülu 3, UltraChat, OpenOrca, Nemotron chat and safety data, the GSM8K training split, and format data it generated}
\mcfield{Own evaluation}{GSM8K test (0.859) and two instruction suites of its own}
\mcfield{Stated limitations}{Exact word counts and punctuation remain hardest; no human study}
\tcblower
\mcstage{Stage 2}{RSIArena/RSIAgent-Jewel-Stage2}
\mcfield{Weights}{LoRA adapter (rank 32)}
\mcfield{Training}{Continued SFT for 80 steps at a low learning rate; the best of about 75 runs}
\mcfield{Data}{A weighted mix with data built from its battles and drills derived from them, and format, reasoning, chat and safety data}
\mcfield{Own evaluation}{Its format suite (0.939) and GSM8K (0.794)}
\mcfield{Stated limitations}{{\color{soft}None stated}}
\end{modelcard}

\end{document}